\documentclass{article}

\usepackage[preprint]{neurips_2026}

\usepackage[utf8]{inputenc} % allow utf-8 input
\usepackage[T1]{fontenc}    % use 8-bit T1 fonts
\usepackage{hyperref}       % hyperlinks
\usepackage{url}            % simple URL typesetting
\usepackage{booktabs}       % professional-quality tables
\usepackage{amsfonts}       % blackboard math symbols
\usepackage{amssymb}        % \checkmark etc.
\usepackage{amsmath} 
\usepackage{nicefrac}       % compact symbols for 1/2, etc.
\usepackage{microtype}      % microtypography
\usepackage{xcolor}         % colors
\usepackage{todonotes}      % TODO notes
\usepackage{graphicx}       % Resize
\usepackage{multirow}       % multirow
\usepackage{booktabs}
\usepackage{wrapfig}
\usepackage{longtable}
\usepackage{siunitx}

\title{ABRA: Agent Benchmark for Radiology Applications\thanks{Project page: \url{https://luab.github.io/abra/}}}

\author{%
  Bulat Maksudov \\
  School of Computing \\
  Dublin City University \\
  Dublin, Ireland \\
  \texttt{bulat.maksudov2@mail.dcu.ie} \\
  \And
  Vladislav Kurenkov \\
  AXXX \\
  \texttt{kurenkov@dunnolab.ai}
  \And
  Kathleen M. Curran \\
  School of Medicine \\
  University College Dublin \\
  Dublin, Ireland \\
  \texttt{kathleen.curran@ucd.ie} \\
  \And
  Alessandra Mileo \\
  School of Computing \\
  Dublin City University \\
  Dublin, Ireland \\
  \texttt{alessandra.mileo@dcu.ie} \\ % Added email if available
}

\begin{document}

\maketitle

\begin{abstract}
Existing medical-agent benchmarks deliver imaging as pre-selected samples, never as an environment the agent must navigate.
We introduce ABRA, a radiology-agent benchmark in which the agent operates an OHIF viewer and an Orthanc DICOM server through twenty-one function-calling tools that span slice navigation, windowing, series selection, pixel-coordinate annotation, and structured reporting.
ABRA contains 655 programmatically generated tasks across three difficulty tiers and eight types (viewer control, metadata QA, vision probe, annotation, longitudinal comparison, BI-RADS reporting, and oracle variants of annotation and BI-RADS reporting), drawn from LIDC-IDRI, Duke Breast Cancer MRI, and NLST New-Lesion LongCT.
Each episode is scored along Planning, Execution, and Outcome~\citep{bluethgen2025agenticsystemsradiologydesign} by task-type-specific automatic scorers.
Ten current models, five closed-weight and five open-weight, reach at least 89\% Execution on real annotation but only 0--25\% Outcome; on the paired oracle variant where a simulated detector supplies the finding, Outcome on the same task reaches 69--100\% across the models evaluated, localising the bottleneck to perception rather than tool orchestration.
Code, task generators, and scorers are released at \url{https://github.com/Luab/ABRA}.
\end{abstract}

\section{Introduction}

Where benchmarks for vision-language models classically deliver the input as a single prompt (an entire image, a concatenated patient record, a pre-curated case vignette), recent agent benchmarks in software, web, and operating-system domains \citep{jimenez2024swebench, zhou2024webarena, xie2024osworld} present the input as an environment the model queries on demand, making observation itself an action the agent chooses to take. Medical imaging is the exception: the multimodal entries among medical agent benchmarks deliver imaging as fixed samples and never as a queryable environment. Radiology offers a sharp test case because the workstation workflow (load a study, scroll slices, adjust windowing, place annotations, compose a report) is already tool-shaped and standardised over DICOM.

Putting the agent inside a live viewer rather than handing it static images turns four behaviours into measurable signals: pixel-coordinate annotation as a scored action, observation cost (which slices, series, and preprocessor the agent chose to query), scroll and windowing as planning steps, and multi-series navigation across longitudinal studies. None of these are exercised when imaging is delivered as preselected samples, and none are exercised by the one prior radiology-specific agent benchmark, RadABench \citep{zheng2024radabench}, whose deliberately symbolic design substitutes placeholder tokens for pixels to upper-bound planning and orchestration with perception out of scope. ABRA is the complementary axis on the same problem: to our knowledge, the first benchmark in which a radiology agent's tool use and visual perception are exercised jointly inside a radiology viewer (full comparison in Table~\ref{tab:benchmark-comparison}).

ABRA wraps the OHIF open-source radiology viewer \citep{Ziegler2020-yv} and an Orthanc DICOM server \citep{Jodogne2018} in a controller that drives a headless browser session and exposes the viewer through a uniform function-calling interface. The tool set is partitioned into four observation categories (metadata, viewer screenshot, DICOM pixel, oracle predictions) and three action classes (navigation, segmentation, reporting).

The benchmark contains 655 tasks generated programmatically from three public TCIA cohorts that together cover chest CT (LIDC-IDRI, NLST New-Lesion LongCT) and breast MRI (Duke Breast Cancer MRI), stratified across three difficulty tiers and eight types ranging from viewer control to end-to-end BI-RADS reporting. Perception-heavy types are released in paired \emph{oracle} and \emph{real} variants of the same task: the oracle variant exposes a simulated-detector tool and no pixel-access tools, while the real variant exposes pixel-access tools and no detector. Each episode is scored along Planning, Execution, and Outcome by task-type-specific automatic scorers.

We evaluate ten current models on ABRA, five closed-weight and five open-weight. Annotation Outcome drops from a 0.69--1.00 oracle range to 0.00--0.25 on the real variant across the roster; the gap is the headline finding and is consistent with perception, not tool orchestration, as the binding constraint. The pattern extends beyond the paired tasks: scores are strong on viewer control and metadata retrieval and weaken progressively as a task focuses on direct pixel interpretation. Current models on ABRA are therefore tool-competent but perception-limited on real DICOMs, which suggests the near-term operating point for clinical assistants is one in which a specialised perception model supplies findings and the LLM orchestrates the workflow around them.

\paragraph{Contributions.}
\begin{itemize}
    \item \textbf{Environment.} An executable radiology-workstation stack (OHIF viewer, Orthanc PACS, preprocessor) exposed to agents through a uniform function-calling interface, with tools partitioned into four observation categories and three action classes covering the workstation workflow.
    \item \textbf{Benchmark.} 655 programmatically synthesised tasks across eight types and three difficulty tiers drawn from three public TCIA datasets covering chest CT and breast MRI, with paired oracle and real variants of the same task that swap access to a simulated detector for access to pixel-level tools.
    \item \textbf{Evaluation.} Automatic Planning, Execution, and Outcome scorers operating on controller trajectory logs, instantiating the framework of \citet{bluethgen2025agenticsystemsradiologydesign} on real episodes.
    \item \textbf{Empirical findings.} Within-pair gaps between oracle and real variants of the same task across current closed- and open-weight models, consistent with perception as the binding constraint and characterised across task types and difficulty tiers.
\end{itemize}

\section{Related Work}
\label{sec:related-work}
A growing body of agent-benchmark research instantiates its tasks inside real, interactive systems rather than static datasets. \citet{xie2024osworld} evaluate agents in real operating systems; \citet{zhou2024webarena} in live Dockerised web stacks; \citet{jimenez2024swebench} against real GitHub repositories with their native test suites; and \citet{mialon2024gaia} on real-world tool use across the live web. In medicine, however, interactive clinical benchmarks remain the exception: most medical agent benchmarks run on simulated clinical workflows rather than the imaging environments radiologists actually use.

To structure the comparison, we adopt the four-tier evaluation framework (Planning, Execution, Outcome, and System-level) proposed by \citet{bluethgen2025agenticsystemsradiologydesign} for agentic radiology systems, and position ABRA against the nine prior medical agent benchmarks they catalogue (Table~\ref{tab:benchmark-comparison}).

\begin{table}[ht]
\centering
\caption{Comparison of ABRA with the nine prior medical agent benchmarks catalogued by \citet{bluethgen2025agenticsystemsradiologydesign} (their Table~2). \emph{Environment}: Static (no interactive env), Chat (dialogue sim), EHR (FHIR/EHR sandbox), Sim (abstract tool simulator), Viewer (live radiology viewer). \emph{Modality}: Text, Multi (text + images). \emph{Tools}: how the agent interacts with the environment.}
\label{tab:benchmark-comparison}
\resizebox{\textwidth}{!}{%
\begin{tabular}{l l l l l}
\toprule
\textbf{Benchmark} & \textbf{Environment} & \textbf{Modality} & \textbf{Tools} & \textbf{Reference} \\
\midrule
\multicolumn{5}{@{}l}{\emph{General medicine / clinical decision making}} \\
CRAFT-MD       & Chat   & Text  & None              & \citet{johri2024craftmd}           \\
AgentClinic    & Chat   & Multi & Tool-call + RAG   & \citet{schmidgall2024agentclinic}  \\
MIMIC-CDM      & EHR    & Text  & Tool-call (3)     & \citet{baniharouni2026lacdm}       \\
MedChain       & Chat   & Multi & None              & \citet{liu2024medchain}            \\
SDBench        & Chat   & Text  & XML actions       & \citet{nori2025sdbench}            \\
MedAgentBench  & EHR    & Text  & 9 FHIR calls      & \citet{jiang2025medagentbench}     \\
MedAgentBoard  & Static & Multi & Python exec       & \citet{zhu2025medagentboard}       \\
MedAgentsBench & Static & Text  & None              & \citet{tang2025medagentsbench}     \\
\midrule
\multicolumn{5}{@{}l}{\emph{Radiology}} \\
RadABench          & Sim             & Text           & XML tool cards             & \citet{zheng2024radabench} \\
\textbf{ABRA}      & \textbf{Viewer} & \textbf{Multi} & \textbf{21 function-calls} & \textbf{this work}         \\
\bottomrule
\end{tabular}%
}
\end{table}

Most prior work targets text-only clinical decision making or medical question-answering and scores primarily task-success, with limited process or cost evaluation. Three benchmarks (AgentClinic, MedChain, MedAgentBoard) include multimodal imaging inputs, but as static image samples rather than through an interactive imaging environment; none exposes a live viewer or a pixel-level action space.

The closest related benchmark is RadABench \citep{zheng2024radabench}, the one other radiology-focused agent benchmark with fine-grained per-step agent evaluation. \citet{zheng2024radabench} deliberately make RadABench a fully symbolic simulator: imaging inputs are placeholder tokens rather than actual pixels, which the authors frame as an ``upper-bound'' measurement of an agent's planning and orchestration capabilities, with perception out of scope. ABRA is the complement on the same axis: a radiology-specific benchmark that exposes an actual imaging stack (OHIF viewer, Orthanc PACS, DICOM pixels), so tool use and visual perception are exercised jointly. To our knowledge, ABRA is the first medical agent benchmark to place the agent inside a live clinical viewer.

\section{ABRA Environment}
\begin{figure}[h]
    \centering    \includegraphics[width=\linewidth]{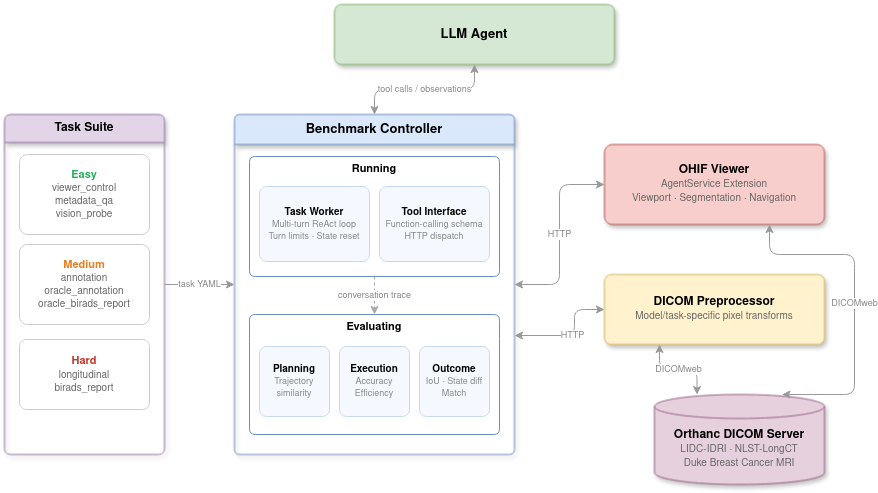}
    \caption{Overview of the ABRA architecture. The benchmark controller draws tasks from the suite, drives an LLM agent through a multi-turn loop, and dispatches its tool calls to a headless OHIF viewer and a DICOM preprocessor, both backed by an Orthanc PACS populated with three TCIA datasets. After the agent terminates, the controller scores the trajectory along Planning, Execution, and Outcome. Episodes run in isolated browser contexts and are trivially parallelisable.}
    \label{fig:architecture}
\end{figure}
In this section we describe the problem formulation, introduce the main components of the environment and observation/action spaces. 
\subsection{Task definition}
\label{sec:task-definition}
Following \citet{xie2024osworld}, we cast an ABRA episode as a goal-conditioned partially observable MDP $(\mathcal{S}, \mathcal{O}, \mathcal{A}, \mathcal{T}, \Omega, r, \rho_0, \mathcal{G})$ specialised to a radiology workstation. $\mathcal{S}$ is the joint OHIF viewer and Orthanc PACS state (loaded studies, viewport geometry, placed annotations, DICOM cache). The observation space $\mathcal{O}$ (Section~\ref{sec:observation-space}) is the set of typed return schemas of the read-only tools, and the action space $\mathcal{A}$ (Section~\ref{sec:action-space}) consists of tool invocations, of which a small terminal subset ends the episode on call. The transition $\mathcal{T}$ and observation function $\Omega$ are the deterministic effect of dispatching tool calls through the controller bridge: only the components of $\mathcal{S}$ named by their return schemas enter the next observation, so observation itself is an action the agent chooses to take. The initial-state distribution $\rho_0$ is degenerate (each task fixes the loaded study, the active series, and the viewport; Section~\ref{sec:environment}). A goal $g \in \mathcal{G}$ is the natural-language instruction together with a structured target (final viewport state, free-text string, segmentation contour, list of longitudinal findings, or BI-RADS report). The reward $r$ is the per-episode composite $S = 0.20\,P + 0.30\,E + 0.50\,O$ defined in Section~\ref{sec:evaluation}, applied once at termination.

\paragraph{Episode dynamics.} At each turn the agent receives the conversation so far and emits a (possibly empty) batch of tool calls from $\mathcal{A}$; the controller dispatches each call, returns its typed result as a separate observation, and re-invokes the agent. Initial context is the instruction plus a snapshot of the reset viewport, with no DICOM tags, pixels, or oracle predictions pre-loaded. Two properties are distinctive. Observations are pull-based, which makes tool selection itself a measurable component of the score (Section~\ref{sec:planning}). Termination is task-shaped, with each task type carrying its own submission action (\texttt{submit\_birads\_report}, \texttt{submit\_longitudinal\_complete}, \texttt{submit\_answer}) rather than a global \texttt{DONE} signal; an episode ends when any of these is invoked, when the agent returns an empty batch with a final text response, or when the agent reaches the task's cap on the number of turns, after which the controller freezes the trajectory and records a final viewport snapshot for the three scorers. If the episode terminates without the task's submission action being invoked, the outcome scorer returns zero for that episode, since no scorable deliverable was produced. Each task fixes this turn cap at generation time; a separate per-response output-token cap is a property of the agent configuration rather than the task, and is reported with the experimental setup in Section~\ref{sec:experimental-evaluation}.

\subsection{Viewer as agent environment}
\label{sec:environment}
ABRA is an executable and resettable environment built around the OHIF open-source radiological viewer \citep{Ziegler2020-yv} paired with an Orthanc DICOM server \citep{Jodogne2018}, mirroring the PACS-plus-viewer stack used in clinical workstations so that the agent workflow (loading a study, scrolling slices, adjusting windowing, placing annotations, writing a report) remains the one radiologists themselves follow. A dedicated OHIF extension surfaces the viewer's internal state (viewport geometry, active series, placed annotations) and the built-in segmentation and measurement modules, while an out-of-process preprocessor converts raw DICOM pixels into model-appropriate PNGs whose coordinate frame is shared with the segmentation actions. ABRA supports two interaction modes over the same underlying tool set: the in-browser chat extension (Figure~\ref{fig:chat_architecture}) executes tools directly inside a user-driven OHIF session, and the benchmark controller drives a headless OHIF instance through Puppeteer and exposes the same tools over an HTTP interface (Figure~\ref{fig:architecture}), so every evaluated agent operates against identical viewer state. Episodes run in isolated browser contexts, which makes them cheap to parallelise and trivially resettable between runs.

\paragraph{Initialization.}
At the start of each episode the controller instructs the Orthanc server to load the study referenced in the task specification, clears any persisted viewer state (active series, windowing presets, placed annotations), and positions the viewport on the default series and slice for that task type. It then assembles the initial context delivered to the agent: the natural-language instruction, the system prompt (Appendix~\ref{apd:task-templates}), and a snapshot of the reset viewport. No pixel data, DICOM tags, or oracle predictions are pre-loaded; every subsequent observation must be fetched through an explicit tool call. Further details are given in Appendix~\ref{apd:initial-state}.

\paragraph{Evaluation.}
Once the agent emits a terminal action (a report submission, a free-text answer, or the longitudinal completion signal), or its step budget is exhausted, the controller freezes the viewer and records three artefacts: the ordered sequence of tool calls, per-call return values and execution metadata, and the final structured deliverable. These feed the three scorers of our evaluation framework (Planning, Execution, and Outcome), whose benchmark-specific formulations are presented in Section~\ref{sec:evaluation}, with full per-task-type breakdowns in Appendix~\ref{apd:evaluation-details}. Browser state is then discarded before the next episode begins.
\subsection{Observation space}
\label{sec:observation-space}
The observation space $\mathcal{O}$ of ABRA is structured as a set of read-only tools that the agent queries on demand. Only the natural-language task instruction and a snapshot of the current viewport state are delivered in the initial prompt; all other information (DICOM tags, pixel content, simulated external-model outputs) must be fetched explicitly through tool calls. This design bounds token consumption on long-horizon episodes and makes tool-call planning a measurable part of the Planning score (Sec.~\ref{sec:planning}). We group observations into four categories: (i)~\emph{Metadata}, six query tools that expose DICOM tags at study, series, and instance granularities, plus the live viewport state and loaded segmentations; (ii)~\emph{Viewer screenshot}, a full-resolution rendering of the OHIF browser UI captured through Puppeteer, intended for inspecting overlays, toolbar state, and already-placed annotations; (iii)~\emph{DICOM pixel}, a preprocessor that routes a requested slice through one of six named pipelines (\texttt{default}, \texttt{lung\_window}, \texttt{soft\_tissue\_window}, \texttt{percentile\_norm}, \texttt{breast\_mri}, \texttt{raw\_uint16}) and returns a PNG whose coordinate frame is shared with the segmentation actions in Sec.~\ref{sec:action-space}; and (iv)~\emph{Oracle predictions}, two tools that simulate calls to external CAD models and return structured findings, which lets us decouple tool-use competence from visual perception. Full tool signatures, return schemas, and preprocessor definitions are given in Appendix~\ref{apd:observation_space}.

\subsection{Action space}
\label{sec:action-space}
Actions in ABRA fall into three classes. \emph{Navigation and display} actions modify the active viewport and therefore change what the agent will next observe. \emph{Segmentation} actions deposit region-of-interest annotations in pixel coordinates; providing circle and bounding-box primitives alongside arbitrary polygons matches the shapes used in typical radiology workflows and keeps the interaction surface narrow without losing the capabilities we want to measure. \emph{Reporting} actions produce the structured deliverable for the task (BI-RADS report, longitudinal finding, free-text answer). Most reporting actions are \emph{terminal}, ending the episode on invocation and handing control to the outcome scorer. The per-episode tool set is scoped by task type (for example, viewer-control tasks see only the navigation class), so every episode evaluates both the agent's ability to select a correct tool and its ability to parameterise it correctly. Pixel coordinates are returned by \texttt{get\_dicom\_image} and accepted by every segmentation action in the same frame, so they can be round-tripped between observations and actions without an explicit transformation. The complete action specification is provided in Appendix~\ref{apd:action_space}.
\begin{wrapfigure}[15]{r}{0.65\textwidth}
    \centering
    \includegraphics[width=0.65\textwidth]{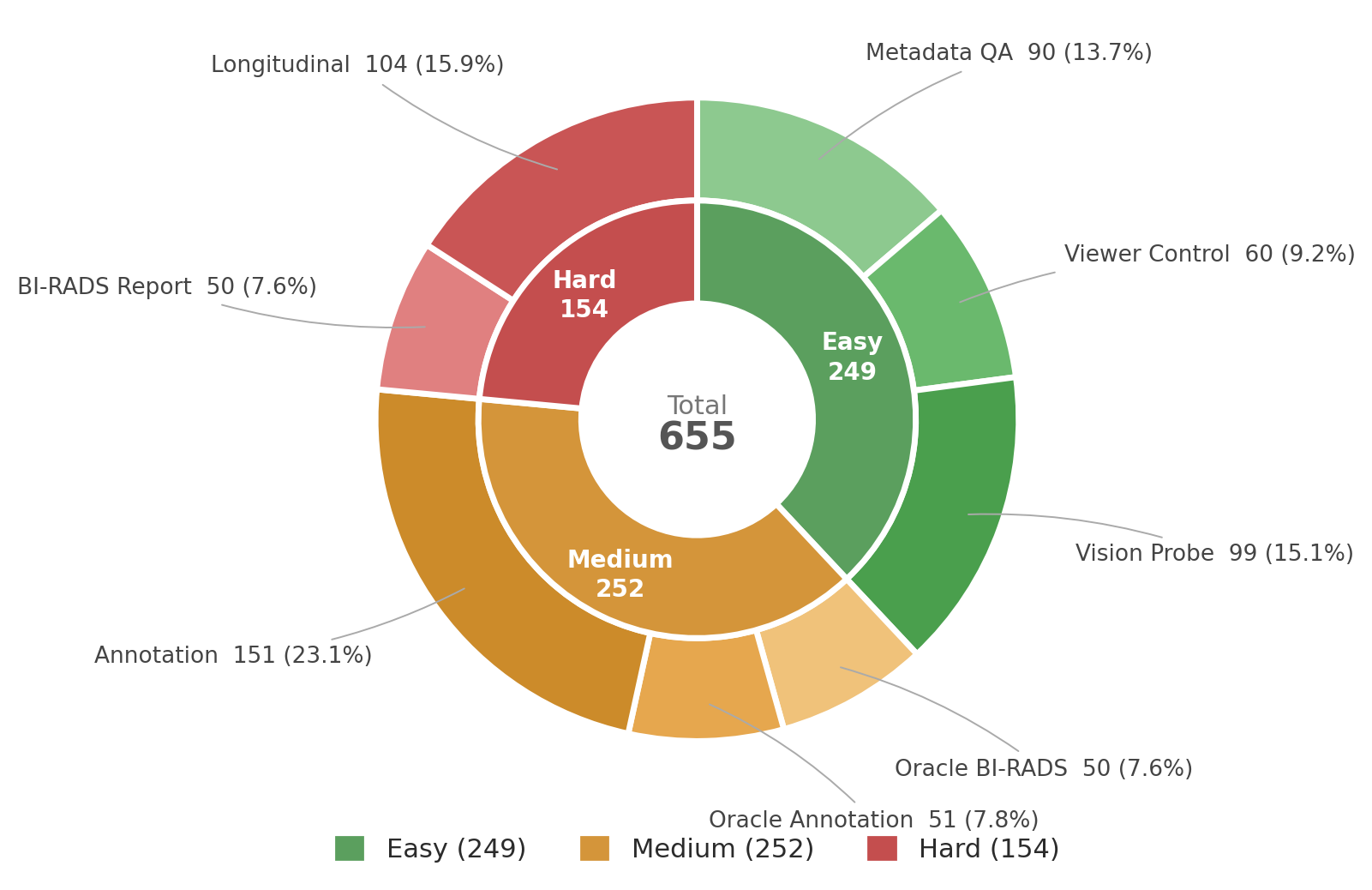}
    \caption{Task distribution by difficulty and type for ABRA.}
    \label{fig:task-distribution}
\end{wrapfigure}

\section{ABRA Benchmark}
\subsection{Task generation}
\label{sec:task-generation}
ABRA tasks are synthesised programmatically from three public TCIA cohorts \citep{clark2013cancer}, stratified over three difficulty tiers, and instantiated from one of eight task types that span the radiologist's workstation workflow. Each generated task contains a natural-language instruction, an initial viewer state, a ground-truth target, and a reference tool-call trajectory used by the planning scorer. We provide 655 tasks across the three source cohorts; their joint distribution over types and difficulties is shown in Figure~\ref{fig:task-distribution}.

\paragraph{Datasets.}
The three datasets were selected to exercise distinct diagnostic scenarios through a shared acquisition pipeline. \emph{LIDC-IDRI} \citep{Armato2011-kc} contributes thoracic CT studies with per-nodule contours from up to four radiologists, aggregated into consensus segmentations using the 50\% volumetric majority-vote convention; these anchor the annotation-heavy middle tier. \emph{Duke Breast Cancer MRI} \citep{DUKE} contributes multi-sequence breast MRI studies with BI-RADS labels and drives the structured-reporting tasks. \emph{NLST New-Lesion LongCT} \citep{NLST-LONGCT} contributes curated baseline-and-follow-up CT pairs with new-lesion annotations and drives the longitudinal comparison tasks. Per-dataset cohort sizes, modality counts, and task contributions are detailed in Appendix~\ref{apd:dataset-selection}.

\paragraph{Difficulty tiers.}
Tasks are partitioned into three tiers whose composition lets us probe different capability profiles. The \emph{Easy} tier contains short viewer-control, metadata, and modality-recognition tasks: either a simple tool call against the viewer or DICOM metadata, or a single-image judgement about the study's modality and appropriate preprocessor. The \emph{Medium} tier splits along a deliberate axis: some tasks require the agent to perceive anatomy directly on CT and place a region-of-interest contour, while the oracle variants hand the model structured findings from a simulated detector and ask it to act on them, separating reasoning and tool orchestration from visual perception. The \emph{Hard} tier targets realistic clinical workflows end-to-end: longitudinal comparison between a baseline and a follow-up CT, and structured BI-RADS reporting on a full multi-sequence breast MRI.

\paragraph{Task types.}
The eight task types are summarised in Table~\ref{tab:task-types}; prompt templates and the per-type tool-availability matrix are given in Appendix~\ref{apd:task-templates}.

\begin{table}[ht]
\centering
\small
\caption{The eight ABRA task types, grouped by difficulty tier.}
\label{tab:task-types}
\begin{tabular}{@{}l l p{8.4cm}@{}}
\toprule
\textbf{Tier} & \textbf{Type} & \textbf{Description} \\
\midrule
Easy   & \texttt{viewer\_control}        & Viewport manipulation: slice navigation, window/level presets, series switching. \\
       & \texttt{metadata\_qa}           & DICOM tag retrieval at study, series, and instance granularity. \\
       & \texttt{vision\_probe}          & Modality recognition and preprocessor selection from a single rendered image. \\
\midrule
Medium & \texttt{annotation}             & Perceive pathology on a CT series and contour it with a segmentation primitive. \\
       & \texttt{oracle\_annotation}     & Same target as \texttt{annotation}, with oracle detector findings supplied. \\
       & \texttt{oracle\_birads\_report} & Compose a BI-RADS report from oracle breast-MRI findings. \\
\midrule
Hard   & \texttt{longitudinal}           & Compare a baseline-and-follow-up CT pair and submit each new lesion with its slice and pixel location. \\
       & \texttt{birads\_report}         & Read a full multi-sequence breast MRI and produce an end-to-end BI-RADS report. \\
\bottomrule
\end{tabular}
\end{table}
\subsection{Tools}
\label{sec:tools}
The twenty-one tools that together form the observation and action spaces were introduced in Sections~\ref{sec:observation-space} and~\ref{sec:action-space}; we briefly describe how they are surfaced to the agent. Each tool is presented as a typed JSON function-calling schema (in the form introduced by OpenAI), and the controller translates every schema invocation into an HTTP request against the benchmark bridge described in Section~\ref{sec:environment}. The per-task-type tool visibility set (detailed in Appendix~\ref{apd:task-templates}) is injected into the agent's initial prompt, so an agent never sees a tool that is irrelevant to its current task.

\subsection{Evaluation}
\label{sec:evaluation}
ABRA scoring instantiates three of the four tiers in the framework that \citet{bluethgen2025agenticsystemsradiologydesign} propose for evaluating agentic radiology systems (their Figure~4): Planning, Execution, and Outcome. Where that work defines the tiers conceptually, we instantiate each with automatic scorers that run on the controller's trajectory logs without any human-in-the-loop review. Every episode yields a composite score $S = 0.20\,P + 0.30\,E + 0.50\,O$, with $P$, $E$, $O$ the Planning, Execution, and Outcome scores in $[0,1]$. The weighting reflects that task success is the end goal, while Planning and Execution interpret \emph{how} success (or failure) was reached. Full formulas for every component are given in Appendix~\ref{apd:evaluation-details}; the subsections below summarise the main design choices.

\paragraph{Planning.}
\label{sec:planning}
We compare the agent's ordered trajectory $T_\text{agent}$ against a reference trajectory $T_\text{ref}$ attached to each task at generation time, collapsing both to unordered tool multisets and scoring F1 between them. Unordered matching was chosen because multiple tool orderings correspond to valid radiological workflows (for example, reviewing metadata before or after scrolling through slices), and penalising order would encode a single canonical sequence. Extraneous calls not present in $T_\text{ref}$ incur a small redundancy penalty (0.05 per extra call, capped at 0.30), which discourages pathological ``call every tool'' behaviour without erasing credit for legitimate exploration.

\textbf{Design choice and limitation.} Reference trajectories are generated programmatically from task templates. This is a best-effort estimate of the reference plan rather than an oracle ground truth. The high Planning and Execution scores on the easy and medium tiers in Table~\ref{tab:abra-results} are consistent with the programmatic references capturing plausible solution routes when the workflow is short and well-constrained, and their divergence on the hard tier marks the regime where independent validation would be most informative. A detailed discussion, together with a plan to collect reference trajectories from real radiologist sessions, is given in Appendix~\ref{apd:future-work}.

\paragraph{Execution.}
\label{sec:execution-main}
Execution assesses the agent step-by-step inside the Reason-Act-Observe loop. Four components are aggregated into a weighted sum $E = 0.40\,A_\text{tool} + 0.20\,Q_\text{param} + 0.25\,E_\text{turn} + 0.15\,R_\text{err}$, where $A_\text{tool}$ is the fraction of tool calls that returned success, $Q_\text{param}$ the semantic quality of the arguments passed to each call, $E_\text{turn}$ turn efficiency relative to the reference trajectory length, and $R_\text{err}$ whether the agent recovered gracefully from a tool error instead of looping. Tool-call success carries the largest weight because failed calls (whether from wrong tool selection or malformed arguments) most directly cascade into bad outcomes. Parameter quality is weighted more lightly because most type-level errors are already caught by the function-calling schema, so what remains to score is semantic quality (selecting an appropriate preprocessor, targeting the correct series UID). Turn efficiency penalises aimless long trajectories; error recovery rewards the rarer but valuable behaviour of turning a tool failure into a correct next action.

\paragraph{Outcome.}
\label{sec:outcome-main}
Outcome evaluates whether the task was completed correctly, independent of how the agent got there. Because the eight task types produce structurally different deliverables (numeric viewer state, free-text answer, segmentation contour, structured BI-RADS report, list of longitudinal findings), we use task-type-specific scorers rather than a single metric, summarised in Table~\ref{tab:outcome-scorers}. Full scorer definitions, their input formats, and the BI-RADS field-weight rationale are given in Appendix~\ref{apd:outcome}.

\begin{table}[ht]
\centering
\small
\caption{ABRA outcome scorers, listed by task type. All scores are normalised to $[0,1]$ so the composite remains bounded.}
\label{tab:outcome-scorers}
\begin{tabular}{@{}l p{4.0cm} p{6.4cm}@{}}
\toprule
\textbf{Scorer} & \textbf{Task type(s)} & \textbf{Definition} \\
\midrule
state-diff      & \texttt{viewer\_control}                                  & Exact match of the final viewport state against the target. \\
exact-match     & \texttt{metadata\_qa}, \texttt{vision\_probe}             & String-level match of the submitted answer. \\
IoU             & \texttt{annotation}, \texttt{oracle\_annotation}          & Intersection-over-union against the consensus contour, divided by the IoU of the best-fit primitive of the agent's chosen shape so a correctly placed circle is not penalised against a polygon. \\
point-distance  & \texttt{longitudinal} (single-lesion)                     & Pixel distance between the submitted centre and the ground-truth lesion centre. \\
multi-finding   & \texttt{longitudinal} (multi-lesion)                      & Detection rate minus $0.1$ per false positive, clamped to $[0,1]$, with the same slice-and-distance match criterion as the single-lesion scorer. \\
BI-RADS-field   & \texttt{birads\_report}, \texttt{oracle\_birads\_report}  & Weighted average of per-field agreement (laterality, category, lesion count, enhancement, quadrant). \\
\bottomrule
\end{tabular}
\end{table}
\section{Experimental evaluation}
\label{sec:experimental-evaluation}

We evaluate ten model checkpoints across two access regimes. Five are vendor-locked frontier APIs (Claude Sonnet 4.6 \citep{anthropic2026sonnet46}; GPT-5.4 and GPT-5.4-nano \citep{singh2026openaigpt5card}; Gemini 3 Flash \citep{google2026gemini3flash}; Gemini 3 Pro \citep{google2025gemini3pro}); five are open-weights checkpoints accessed via Ollama Cloud (Gemma 4 \citep{google2026gemma4}; Qwen 3.5 \citep{qwen35blog}; Ministral 3 14B \citep{liu2026ministral3}; Mistral Large 3 \citep{mistral2025large3}; Kimi K2.5 \citep{kimiteam2026kimik25visualagentic}). Each task's turn cap is fixed at generation time (Section~\ref{sec:task-definition}); we additionally cap output at 20{,}048 tokens per response, decode at temperature~0 with provider-default reasoning effort (medium across all evaluated models), and report a single run per (model, task) pair. Anthropic prompt caching is invoked explicitly on the system block and a trailing breakpoint over the growing conversation, while OpenAI and Gemini rely on the providers' automatic prefix caching. Tasks that exit on the turn cap without invoking the relevant submit action receive Outcome\,$=\,0$.
\sisetup{
  table-format=1.2,
  table-number-alignment=center,
  detect-weight=true,
  detect-inline-weight=math
}

\begin{table}[ht]
\centering
\caption{ABRA evaluation across difficulty tiers. Per-tier scores are Planning (P), Execution (E), Outcome (O), and composite average $S = 0.20P + 0.30E + 0.50O$. The final column reports the $n$-weighted average across all eight task types. Higher is better.}
\label{tab:abra-results}
\small
\renewcommand{\arraystretch}{1.15}
\setlength{\tabcolsep}{4pt}

\begin{tabular}{
l
S S S S
S S S S
S S S S
S
}
\toprule
& \multicolumn{4}{c}{\textbf{Easy}} 
& \multicolumn{4}{c}{\textbf{Medium}} 
& \multicolumn{4}{c}{\textbf{Hard}} 
& \textbf{Overall} \\
\cmidrule(lr){2-5} \cmidrule(lr){6-9} \cmidrule(lr){10-13}
\textbf{Model}
& {P} & {E} & {O} & {\textbf{Avg}}
& {P} & {E} & {O} & {\textbf{Avg}}
& {P} & {E} & {O} & {\textbf{Avg}}
& {} \\
\midrule

Claude Sonnet 4.6
& 0.93 & 0.99 & 0.86 & 0.91
& 0.78 & {\bfseries 0.99} & 0.41 & 0.66
& 0.20 & 0.98 & {\bfseries 0.21} & 0.44
& {\bfseries 0.70} \\

GPT-5.4
& 0.83 & 0.99 & 0.88 & 0.91
& {\bfseries 0.82} & {\bfseries 0.99} & 0.42 & {\bfseries 0.67}
& 0.31 & 0.95 & 0.11 & 0.40
& {\bfseries 0.70} \\

Qwen 3.5
& 0.88 & {\bfseries 1.00} & {\bfseries 0.91} & {\bfseries 0.93}
& 0.76 & 0.93 & 0.39 & 0.63
& 0.49 & {\bfseries 0.99} & 0.11 & 0.45
& {\bfseries 0.70} \\

GPT-5.4-nano
& {\bfseries 0.95} & 0.99 & 0.87 & 0.92
& 0.71 & {\bfseries 0.99} & 0.39 & 0.64
& {\bfseries 0.56} & 0.97 & 0.04 & 0.42
& 0.69 \\

Gemma 4
& 0.88 & {\bfseries 1.00} & 0.87 & 0.91
& 0.75 & 0.93 & 0.43 & 0.64
& 0.38 & 0.93 & 0.02 & 0.37
& 0.68 \\

Mistral Large 3
& 0.90 & 0.98 & 0.69 & 0.82
& 0.80 & {\bfseries 0.99} & 0.38 & 0.65
& 0.51 & 0.90 & 0.15 & 0.45
& 0.67 \\

Gemini 3 Flash
& 0.61 & 0.93 & 0.88 & 0.84
& 0.58 & 0.95 & {\bfseries 0.53} & {\bfseries 0.67}
& 0.37 & 0.86 & 0.06 & 0.36
& 0.66 \\

Ministral 3 (14B)
& 0.92 & 0.99 & 0.68 & 0.82
& 0.78 & 0.98 & 0.38 & 0.64
& 0.43 & 0.87 & 0.12 & 0.41
& 0.66 \\

Gemini 3 Pro
& 0.62 & 0.94 & 0.79 & 0.80
& 0.62 & 0.98 & 0.35 & 0.59
& 0.33 & 0.96 & 0.16 & 0.44
& 0.64 \\

Kimi K2.5
& 0.73 & 0.89 & 0.75 & 0.79
& 0.68 & 0.98 & 0.37 & 0.61
& 0.44 & 0.98 & 0.14 & {\bfseries 0.46}
& 0.64 \\

\midrule
\multicolumn{14}{p{\linewidth}}{\footnotesize\textit{API snapshots:} GPT-5.4 (\texttt{gpt-5.4-2026-03-05}), GPT-5.4-nano (\texttt{gpt-5.4-nano-2026-03-17}), Gemini 3 Flash (\texttt{google/gemini-3-flash-preview-20251217}), Gemini 3 Pro (\texttt{google/gemini-3.1-pro-preview-20260219}).}
\\
\bottomrule
\end{tabular}
\end{table}

Table~\ref{tab:execution-budget} reports per-model token consumption and serialised runtime across the full benchmark, broken down by difficulty tier. Per-task-type breakdowns with prompt-cache hit ratios are given in Appendix~\ref{apd:token-usage}.

We observe that Outcome scores broadly follow the difficulty grading: Easy-tier Outcome stays in the upper range, Medium drops substantially, and Hard converges to near-floor values. Both the Medium and Hard tier averages, however, mask a non-uniform contribution from their constituent task types. On Medium, annotation without oracle conditioning largely fails, while the oracle-conditioned variants mostly succeed and account for most of the achieved Medium score. On Hard, the achieved score is dominated by BI-RADS reporting, which admits competitive scores from prior-based guessing on a well-known categorical distribution and earns partial credit on off-by-one ordinal errors; longitudinal new-lesion detection admits no comparable shortcut (per-task-type breakdown in Appendix~\ref{apd:per-task-results}). The Execution sub-score, by contrast, sits near its ceiling on every tier and for most models (Execution $\approx 1.0$): tool calls succeed, function arguments are correctly typed, and trajectories close within the per-task turn budget at near-unit rates.

\subsection{Analysis}
\label{sec:analysis}

\begin{table}[h]
\centering
\small
\caption{Outcome and overall Avg on the four paired (oracle vs \emph{real}) task variants: annotation and BI-RADS reporting.}
\label{tab:oracle-vs-real}
\begin{tabular}{l cc cc cc cc}
\toprule
\multirow{3}{*}{\textbf{Model}} & \multicolumn{4}{c}{\textbf{Oracle}} & \multicolumn{4}{c}{\textbf{Real}} \\
\cmidrule(lr){2-5} \cmidrule(lr){6-9}
 & \multicolumn{2}{c}{Annotation} & \multicolumn{2}{c}{BI-RADS} & \multicolumn{2}{c}{Annotation} & \multicolumn{2}{c}{BI-RADS} \\
\cmidrule(lr){2-3} \cmidrule(lr){4-5} \cmidrule(lr){6-7} \cmidrule(lr){8-9}
 & Out $\uparrow$ & \textbf{Avg $\uparrow$} & Out $\uparrow$ & \textbf{Avg $\uparrow$} & Out $\uparrow$ & \textbf{Avg $\uparrow$} & Out $\uparrow$ & \textbf{Avg $\uparrow$} \\
\midrule
Claude Sonnet 4.6 & \textbf{1.00} & \textbf{0.98} & \textbf{1.00} & \textbf{0.96} & 0.02 & 0.45 & \textbf{0.64} & \textbf{0.73} \\
GPT-5.4           & 0.98 & 0.97 & \textbf{1.00} & \textbf{0.96} & 0.03 & 0.47 & 0.32 & 0.53 \\
Mistral Large 3   & 0.91 & 0.93 & \textbf{1.00} & \textbf{0.96} & 0.00 & 0.45 & 0.47 & 0.60 \\
Gemini 3 Flash    & 0.90 & 0.82 & \textbf{1.00} & 0.93 & \textbf{0.25} & \textbf{0.53} & 0.18 & 0.41 \\
Qwen 3.5          & 0.95 & 0.94 & \textbf{1.00} & \textbf{0.96} & 0.00 & 0.42 & 0.35 & 0.59 \\
Kimi K2.5         & 0.84 & 0.80 & 0.94 & 0.93 & 0.02 & 0.45 & 0.42 & 0.62 \\
Ministral 3 (14B) & 0.90 & 0.91 & \textbf{1.00} & \textbf{0.96} & 0.00 & 0.45 & 0.35 & 0.46 \\
GPT-5.4-nano      & 0.96 & 0.94 & \textbf{1.00} & \textbf{0.96} & 0.00 & 0.42 & 0.12 & 0.45 \\
Gemini 3 Pro      & 0.69 & 0.73 & 0.58 & 0.71 & 0.16 & 0.51 & 0.50 & 0.65 \\
Gemma 4           & 0.88 & 0.87 & \textbf{1.00} & \textbf{0.96} & 0.08 & 0.46 & 0.06 & 0.33 \\
\bottomrule
\end{tabular}
\end{table}

\paragraph{Vision is the dominant bottleneck.} Across task types, the location and depth of failures point to vision as the dominant constraint. Annotation without oracle conditioning sits near the floor of the Outcome scale on every model in the roster, and the residual Outcome on Hard concentrates in the BI-RADS task type where prior-based guessing is available; the longitudinal new-lesion task, which requires actually localising imaging findings on follow-up volumes, remains at near-zero across the roster (per-task-type breakdown in Appendix~\ref{apd:per-task-results}). One observation in the opposite direction is worth noting: the Gemini family (both Flash and Pro) produces the only annotation results in the roster that score meaningfully above chance on the more pronounced LIDC-IDRI nodules.

\paragraph{Tool-call mechanics.} Execution sub-scores sit at the top of their range across the entire roster (Table~\ref{tab:abra-results}); Planning is high on Easy and Medium but drops on Hard, a limitation of the synthetic reference trajectories (Appendix~\ref{apd:future-work}). This highlights that the capability gap of current models is mostly clinical: agents drive the viewer, query DICOM tags, and invoke submission actions along the trajectories the benchmark expects, with rough congruence to the reference trajectories used by the Planning scorer. At the same time, models occasionally omit findings on multi-lesion longitudinal tasks, and the smaller checkpoints sometimes alter oracle annotations that should pass through unchanged (oracle-alteration rates per model in Table~\ref{tab:oracle-behaviour}). Clinical reliability, rather than tool-call mechanics, appears to be the tighter constraint on Outcome.

\section{Conclusion}
\label{sec:conclusion}

ABRA is an agent benchmark that places the model inside a live radiology workstation: an OHIF viewer and Orthanc PACS driven through a controller bridge that exposes function-calling tools spanning observation, navigation, segmentation, and structured reporting. On top of this environment we release 655 programmatically synthesised tasks across eight types and three difficulty tiers, drawn from three public TCIA cohorts, with paired oracle and real variants that isolate visual perception from tool orchestration; trajectories are scored along Planning, Execution, and Outcome \citep{bluethgen2025agenticsystemsradiologydesign}. Empirically, frontier models drive the viewer and compose structured deliverables at near-ceiling Execution, but Outcome on real annotation variants falls to 0.00--0.25 while paired oracle variants recover to 0.69--1.00, localising the gap to perception rather than tool-call mechanics on the current generation. We position ABRA as a measurement substrate for that gap rather than a verdict on clinical readiness; benchmark limitations and the future directions they motivate are discussed in Appendix~\ref{apd:future-work}.

\begin{ack}
\end{ack}

%\section*{References}
\medskip

{
\small

\bibliographystyle{plainnat} % Or plainnat, unsrt, etc. depending on your preference
\bibliography{references} % This must match the name of your .bib file without the extension

%%%%%%%%%%%%%%%%%%%%%%%%%%%%%%%%%%%%%%%%%%%%%%%%%%%%%%%%%%%%

\appendix

\section{Details of ABRA Environment}

\subsection{Environment Infrastructure}
\label{apd:env-infrastructure}

The ABRA environment is composed of three services that together reproduce a clinical workstation: an Orthanc DICOM server acting as the PACS, an OHIF viewer (v3.9.0) fronted by a custom extension that surfaces internal OHIF tools for model usage, and a Python preprocessing sidecar that handles all DICOM pixel work.

\paragraph{Preprocessor sidecar.}
DICOM pixel work is handled outside the viewer. The sidecar pulls the requested instance from Orthanc via WADO-RS, runs one of the six named pipelines (Section~\ref{sec:observation-space}) that cover modality-appropriate windowing and model-specific input formatting, and returns a base64 PNG anchored to the coordinate frame used by the segmentation actions. Decoupling preprocessing from the viewer keeps pixel transformations deterministic and independent of viewer rendering state.

\paragraph{User-facing interface.}
The viewer includes a chat panel extension (Figure~\ref{fig:chat_architecture}) that surfaces the same tool set as the benchmark, connected to a model endpoint of the user's choice.

\paragraph{Benchmark automation.}
For automated evaluation, a separate Node bridge drives a headless instance of this environment, mapping each HTTP request from the benchmark controller to a typed entry point inside the OHIF page; the same tool surface is therefore exercised whether a user or an LLM is at the other end. A parallel evaluation mode replicates the viewer and bridge into multiple independent instances running against the shared Orthanc PACS, with the controller dispatching one task to each free instance.

\begin{figure}[h]
    \centering
    \includegraphics[width=\linewidth]{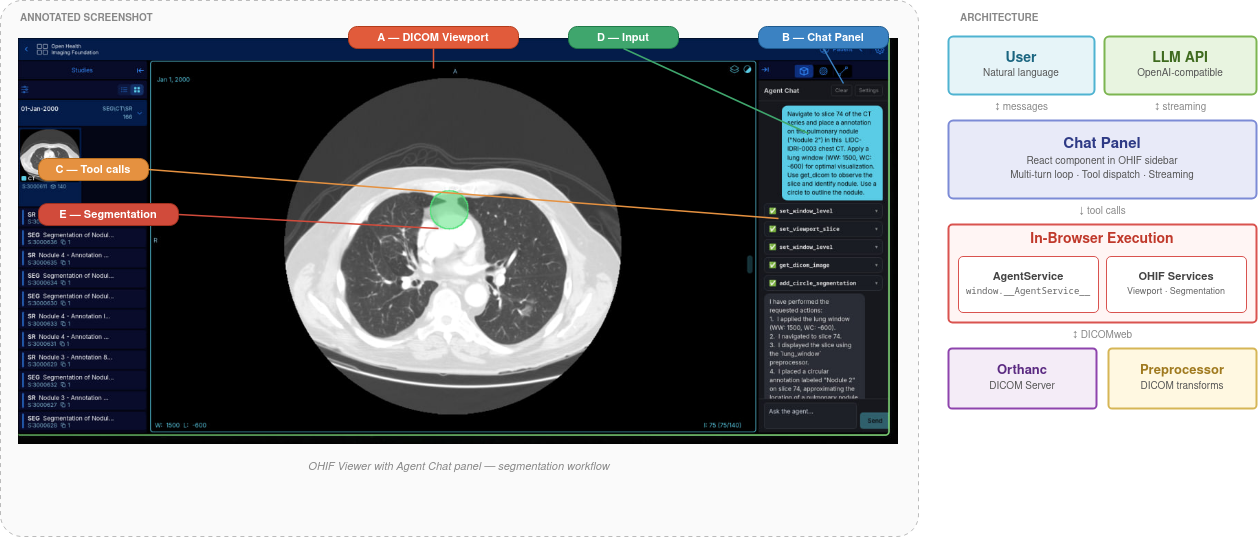}
\caption{Agent execution path inside OHIF. A user prompt (D)
  submitted via the embedded chat panel (B) drives streamed tool calls
  (C) executed in-browser against the viewport (A) and segmentation
  layer (E); DICOM data is stored in Orthanc, and pixel observations
  reach the agent through a model-specific preprocessor.}    \label{fig:chat_architecture}
\end{figure}

\subsection{Observation Space}
\label{apd:observation_space}

ABRA exposes observations as a set of read-only tools the agent invokes on demand. Only the natural-language task instruction is delivered in the initial prompt; all other state (such as study headers, slice pixels, viewport settings, and external-model outputs) must be queried explicitly, which bounds token usage on long-horizon episodes and makes tool-call planning an important part of scoring. All pixel observations share a single coordinate convention: the $(x,y)$ frame returned by \texttt{get\_dicom\_image} matches the frame expected by every segmentation action in \ref{apd:action_space}. We implement four observation categories: \textbf{metadata}, \textbf{viewer screenshot}, \textbf{DICOM pixel}, and \textbf{oracle predictions}.

\subsubsection{Metadata}
Six read-only query tools return DICOM tags at three granularities, the current viewport state, and loaded annotations:

\begin{itemize}
  \item \texttt{get\_study\_metadata(study\_uid)}: study-level tags (\texttt{StudyDate}, \texttt{PatientID}, \texttt{PatientName}, \texttt{Modality}, \texttt{StudyDescription}) plus a summary list of series (UID, description, modality, series number, instance count).
  \item \texttt{get\_study\_series(study\_uid)}: same series list with up to three sample instances per series, enabling \texttt{SOPInstanceUID} selection without paging through the full study.
  \item \texttt{get\_series\_metadata(series\_uid)}: geometry (\texttt{SliceThickness}, \texttt{PixelSpacing}, \texttt{ImageOrientationPatient}, \texttt{Rows}, \texttt{Columns}) and body-part information for a single series.
  \item \texttt{get\_instance\_metadata(study\_uid, series\_uid, sop\_uid)}: full DICOM tag set for a single SOP instance, including windowing defaults (\texttt{WindowCenter}, \texttt{WindowWidth}), rescale parameters (\texttt{RescaleSlope}, \texttt{RescaleIntercept}), and geometric positioning (\texttt{ImagePositionPatient}).
  \item \texttt{get\_viewport\_state()}: current viewport: slice index, total image count, window width/centre, zoom, active \texttt{SeriesInstanceUID}, and loaded \texttt{displaySetInstanceUIDs}.
  \item \texttt{list\_segmentations()}: all segmentations currently loaded in the viewer, with identifiers, labels, and per-segment details.
\end{itemize}
\subsubsection{Viewer Screenshot}
\texttt{get\_viewer\_screenshot()} captures the full OHIF browser UI as a base64-encoded PNG through Puppeteer. The screenshot includes all on-screen overlays (hanging protocol layout, toolbar state, and any annotations the agent has already placed), allowing the model to inspect the UI directly. For medical image interpretation we provide \texttt{get\_dicom\_image} instead, which allows for model-specific preprocessing and high-fidelity input for vision-based tasks.
\subsubsection{DICOM Pixel}
DICOM pixel values are 12--16-bit and modality-dependent, so a raw array is rarely usable by a vision--language model without windowing and channel rearrangement. \texttt{get\_dicom\_image(study\_uid, series\_uid, slice\_index, preprocessor)} routes a requested slice through a named preprocessing pipeline in the \texttt{preprocessor} sidecar and returns a base64 PNG whose pixel dimensions match the coordinate frame expected by the segmentation actions. We implement six pipelines: \texttt{default} (modality-aware default windowing), \texttt{lung\_window}, \texttt{soft\_tissue\_window}, \texttt{percentile\_norm}, \texttt{breast\_mri}, and \texttt{raw\_uint16}. Each task description names the expected preprocessor, and the execution scorer penalises modality-incompatible choices (e.g., a CT preset applied to an MR study).
\subsubsection{Oracle Predictions}
To decouple tool-use competence from visual perception, we expose two oracle tools that simulate calls to external CAD models:

\begin{itemize}
  \item \texttt{query\_pathology\_model(series\_uid, slice\_index?)} simulates a nodule detector on a CT series. Without \texttt{slice\_index} it returns an overview of all detected findings (labels, slice ranges, confidence scores, representative slices). With a specific \texttt{slice\_index} it returns the polygon contour on that slice in pixel coordinates, directly compatible with \texttt{add\_polygon\_segmentation}.
  \item \texttt{query\_birads\_model(series\_uid)} simulates a breast-MRI CAD model and returns structured BI-RADS findings: laterality, lesion count, BI-RADS assessment category, enhancement status, and per-lesion quadrant where available.
\end{itemize}
\subsection{Action Space}
\label{apd:action_space}

Actions in ABRA fall into three classes: \textbf{navigation and display}, \textbf{segmentation}, and \textbf{reporting}. A subset of actions is \emph{terminal}: invoking one ends the episode and hands control to the outcome scorer. Task-type-specific tool sets restrict which classes an agent can use (e.g., viewer-control tasks see only navigation tools). Segmentation and longitudinal actions take pixel-space coordinates in the frame returned by \texttt{get\_dicom\_image}.

\subsubsection{Navigation and Display}
Four tools modify the active viewport; their effect is observed via \texttt{get\_viewport\_state} or \texttt{get\_viewer\_screenshot}.

\begin{itemize}
  \item \texttt{set\_viewport\_slice(slice\_index)} navigates to a 0-based slice index.
  \item \texttt{set\_window\_level(window\_width, window\_center)} sets the display window width and centre; the values are passed straight to OHIF's windowing controls, so the accepted range depends on the modality of the loaded series.
  \item \texttt{set\_zoom(scale)} sets the viewport zoom factor; smaller values zoom in.
  \item \texttt{select\_series(series\_uid)} switches the active viewport to a different series within the loaded study.
\end{itemize}
\subsubsection{Segmentation}
ABRA exposes three annotation primitives that cover the ROI shapes used in typical radiology workflows. Providing circle and bounding-box tools alongside the general polygon simplifies the interaction surface while still capturing the capabilities we want to measure: a VLM's ability to perceive the relevant structures and apply medical-domain reasoning to image coordinates. The IoU scorer normalises the agent's IoU against the best-fit shape of the chosen type, so a correctly placed circle or rectangle is not penalised relative to a polygon.

\begin{itemize}
  \item \texttt{add\_circle\_segmentation(label, slice\_index, center, radius)} places a circle in pixel coordinates.
  \item \texttt{add\_rectangle\_segmentation(label, slice\_index, top\_left, bottom\_right)} places an axis-aligned bounding box.
  \item \texttt{add\_polygon\_segmentation(label, slice\_index, points)} places an arbitrary polygon, points given as $[[x_1, y_1], \ldots]$.
\end{itemize}
\subsubsection{Reporting}
Reporting tools produce the structured deliverable for tasks that require one. Some are terminal and end the episode when invoked; others are called multiple times per task, accumulating annotations until a terminal tool is invoked.

\begin{itemize}
  \item \texttt{submit\_birads\_report(laterality, lesion\_count, birads\_category, enhancement\_present, findings?, recommendation?)} submits a structured breast-MRI report. The optional \texttt{findings} array holds per-lesion fields (\texttt{location\_quadrant}, \texttt{type} $\in$ \{\texttt{mass}, \texttt{non\_mass}, \texttt{focus}\}, \texttt{size\_mm}, \texttt{shape}, \texttt{margin}, \texttt{enhancement}) following the BI-RADS MRI lexicon.
  \item \texttt{submit\_longitudinal\_finding(finding\_type, slice\_index, location, description?)} logs one longitudinal finding, with \texttt{finding\_type} $\in$ \{\texttt{new\_lesion}, \texttt{size\_change}, \texttt{no\_change}\} and \texttt{location} in follow-up pixel coordinates. Unlike the other reporting tools, this one is \emph{not} terminal and is called once per finding.
  \item \texttt{submit\_longitudinal\_complete(summary?)} terminates a longitudinal comparison task after all findings have been submitted.
  \item \texttt{submit\_answer(answer)} submits a free-text answer for metadata-QA tasks.
\end{itemize}
\section{Details of ABRA Benchmark}

\subsection{Dataset selection}
\label{apd:dataset-selection}

ABRA draws its studies from three public collections, all accessed through TCIA \citep{clark2013cancer}. Together they span the three most common radiology modalities (CT, DX/CR, MR) and two annotation schemas carried by DICOM itself: SEG volumes for pixel-level contours and SR objects for structured reports. Table~\ref{tab:datasets} summarises the resulting cohort after local filtering, and Table~\ref{tab:dataset-contribution} gives the corresponding task-type contributions.

\begin{table}[ht]
\centering
\small
\caption{Datasets used by ABRA after local filtering. \textit{Studies} counts distinct \texttt{StudyInstanceUID}s; \textit{Series} counts individual DICOM series including derived objects (SEG for segmentation, SR for structured reports). Parenthesised numbers in the modality column give per-modality series counts.}
\label{tab:datasets}
\setlength{\tabcolsep}{3pt}
\resizebox{\textwidth}{!}{%
\begin{tabular}{@{}l l rrr l l l@{}}
\toprule
\textbf{Dataset} & \textbf{Domain} & \textbf{Patients} & \textbf{Studies} & \textbf{Series} & \textbf{Modalities (series)} & \textbf{Reference} & \textbf{License} \\
\midrule
LIDC-IDRI              & Thoracic CT              & 10  & 20  & 142 & CT (10), DX (8), CR (2), SEG (61), SR (61) & \citet{Armato2011-kc} & CC BY 3.0    \\
Duke Breast Cancer MRI & Bilateral breast MRI     & 50  & 50  & 278 & MR (270), SEG (8)                          & \citet{DUKE}          & CC BY-NC 4.0 \\
NLST New-Lesion LongCT & Longitudinal thoracic CT & 121 & 246 & 246 & CT (246)                                   & \citet{NLST-LONGCT}   & CC BY 4.0    \\
\bottomrule
\end{tabular}%
}
\end{table}

\begin{table}[ht]
\centering
\small
\caption{Contribution of each source dataset to the 655 ABRA tasks, with task-type breakdown. The tasks-per-study ratio (rightmost column) is a measure of annotation density: LIDC-IDRI's small study count is offset by the presence of multiple expert-consensus nodules per CT and by rich supplementary series (DX/CR radiographs, SEG, SR) on every study, which together support five different task types from only 20 studies. Duke and NLST tasks are drawn far less densely, reflecting single-modality content with at most one reporting target or one longitudinal finding per case.}
\label{tab:dataset-contribution}
\begin{tabular}{@{}l r p{6.4cm} r r@{}}
\toprule
\textbf{Dataset} & \textbf{Studies} & \textbf{Task types sourced} & \textbf{Tasks} & \textbf{Tasks / study} \\
\midrule
LIDC-IDRI      & 20  & \texttt{viewer\_control}, \texttt{metadata\_qa}, \texttt{annotation}, \texttt{oracle\_annotation}, \texttt{vision\_probe}     & 326 & 16.3 \\
Duke Breast    & 50  & \texttt{oracle\_birads\_report}, \texttt{birads\_report}, \texttt{vision\_probe}                                            & 135 & 2.7  \\
NLST LongCT    & 246 & \texttt{metadata\_qa}, \texttt{longitudinal}, \texttt{vision\_probe}                                                        & 194 & 0.8  \\
\midrule
\textbf{Total} & \textbf{316} & & \textbf{655} & \\
\bottomrule
\end{tabular}
\end{table}

\paragraph{LIDC-IDRI.} LIDC-IDRI is used for every task type that touches thoracic CT: the viewer-control and metadata-QA tiers, the annotation and oracle-annotation tasks, and a subset of the image-only vision probes. For each of 10 patients we retain a primary CT study together with its accompanying chest radiograph study, giving 20 DICOM studies and 142 series in total (10 CT volumes, 8 DX and 2 CR radiographs, 61 SEG objects, and 61 SR reports). Nodules in LIDC-IDRI are contoured independently by up to four thoracic radiologists per patient, and each reader's contour is released as a separate DICOM SEG object, which accounts for the SEG and SR series counts. To obtain a single reference volume per nodule we apply a volumetric 50\% majority-vote consensus, following the convention adopted in the LIDC-IDRI community: per-reader three-dimensional masks are zero-padded over the union $z$-range of all readers who marked the same nodule, averaged, and thresholded at $\geq 0.5$, so that a slice marked by only one of three readers is discarded rather than passing through as a single-reader record. After consensus, the retained subset contains 22 distinct nodules across 9 patients (one patient has no nodule that clears the consensus threshold). These consensus volumes serve as the reference for the IoU-based outcome scorer in annotation tasks. Although the retained LIDC-IDRI cohort is small in patient count, the combination of multi-reader consensus nodules, supplementary radiograph studies, and DICOM-native derived objects yields 326 tasks (16.3 per study), covering five of the eight task types in the benchmark.

\paragraph{Duke Breast Cancer MRI.} The Duke cohort supplies bilateral breast MRI studies with clinical BI-RADS metadata, and is the basis for both BI-RADS reporting task types (\texttt{oracle\_birads\_report} and \texttt{birads\_report}). We retain 50 patients (50 studies, 278 series) whose dynamic contrast-enhanced (DCE) acquisitions are complete. A typical study contains a pre-contrast axial sequence, four post-contrast DCE passes, and a contrast-enhanced T1-weighted series, giving the five to seven MR series visible to the agent at task time. DICOM SEG segmentations are available for a subset of lesions but are not required by the reporting task formulation. Structured BI-RADS attributes (laterality, lesion count, BI-RADS assessment category, enhancement status) are taken from the clinical spreadsheets released with the collection.

\paragraph{NLST New-Lesion LongCT.} NLST-LongCT provides baseline and follow-up low-dose chest CT pairs with expert-annotated new-lesion coordinates, and is the sole source of \texttt{longitudinal} tasks. We ingest 121 participants and 246 CT studies, comprising 125 baseline/follow-up pairs as released by NLST. The released collection partitions into 90 pairs with a single new lesion, 14 with multiple new lesions (up to six), and 21 with no new lesion (released as longitudinal negatives by NLST but not used in ABRA, since the longitudinal task formulation requires a lesion to localise). After excluding one pair with a malformed annotation during task construction, 124 are retained for longitudinal task generation. The curated annotations record 129 new-lesion points across the lesion-bearing pairs. NLST also contributes the \texttt{metadata\_qa} templates that query longitudinal DICOM tags (study-date interval and follow-up slice count) and a share of the modality and preprocessing vision probes.

\subsection{Task generation templates}
\label{apd:task-templates}

ABRA tasks are produced by deterministic  generators. Each generator instantiates a fixed text template with fields drawn from the dataset and per-study metadata (patient ID, study UID, series UID, slice ranges, ground-truth labels). The resulting \texttt{task\_description} string is written to the task YAML and later appended to the agent system prompt at runtime. Placeholders in Table \ref{tab:task-templates} are written in the form \texttt{\{field\}} exactly as they appear in the generator source.

\paragraph{Agent system prompt.} The runtime system prompt is assembled during benchmark runtime. The preamble is fixed; the study-context block is included whenever a \texttt{study\_uid} is set on the task; the viewer-state block is included; and the task description is appended last. The viewer-state JSON contains a focused subset of the viewport state (\texttt{sliceIndex}, \texttt{totalImages}, \texttt{windowWidth}, \texttt{windowCenter}, \texttt{zoom}, \texttt{seriesInstanceUID}, \texttt{displaySetInstanceUIDs}).

\begin{figure}[h]
\centering
\begin{minipage}{0.95\linewidth}
\begin{small}
\begin{verbatim}
You are a radiology AI agent operating inside a medical imaging viewer.
Use the available tools to complete the task described below. Be precise
and efficient -- use only the tools necessary to complete the task. All
coordinates are in pixel space.

Study context:
- StudyInstanceUID: {study_uid}
- SeriesInstanceUID (loaded): {initial_series_uid}

Current viewer state:
{viewer_state_json}

Task: {task_description}
\end{verbatim}
\end{small}
\end{minipage}
\caption{Agent system prompt template. \texttt{\{task\_description\}} is the per-task string from Table \ref{tab:task-templates}; \texttt{\{viewer\_state\_json\}} is a JSON dump of the current OHIF viewport state.}
\label{fig:system-prompt}
\end{figure}

\paragraph{Per-task templates.} Table \ref{tab:task-templates} lists every \texttt{task\_description} template used by the benchmark, grouped first by difficulty tier and then by task type. Near-duplicate probes that differ only in pixel-preprocessing pipeline (such as the \emph{normal} and \emph{noise} variants of a vision-probe classifier) share the same template and are collapsed into a single row. A literal \texttt{\textbackslash n} denotes a newline in the rendered prompt.

\begin{small}
\begin{longtable}{@{}p{2.6cm}p{9.6cm}@{}}
\caption{Task-description templates used by the ABRA generators, grouped by difficulty tier and task type. Placeholders \texttt{\{field\}} are instantiated at generation time from study, series, and annotation metadata.}\label{tab:task-templates}\\
\toprule
\textbf{Task ID prefix} & \textbf{Template} \\
\midrule
\endfirsthead
\multicolumn{2}{@{}l}{\emph{Table \ref{tab:task-templates} continued from previous page.}}\\
\toprule
\textbf{Task ID prefix} & \textbf{Template} \\
\midrule
\endhead
\bottomrule
\endfoot

\multicolumn{2}{@{}l}{\textbf{Easy} $\cdot$ \texttt{viewer\_control}}\\
\midrule
\texttt{t1\_slice} & \ttfamily Navigate to slice \{target\} of the current CT series. \\
\addlinespace
\texttt{t1\_wl\_*} & \ttfamily Set the window width to \{preset.ww\} and window center to \{preset.wc\} for a standard \{preset.label\} on this \{patient\_id\} chest CT. \\
\addlinespace
\texttt{t1\_slice\_wl} & \ttfamily Navigate to slice \{target\} and apply a bone window (window width 2500, window center 480) on this \{patient\_id\} CT. \\
\addlinespace
\texttt{t1\_series} & \ttfamily The current viewport shows a CT series from \{patient\_id\}. First query the study metadata to discover available series, then select the \{target.modality\} series (``\{target.description\}''). \\
\midrule
\multicolumn{2}{@{}l}{\textbf{Easy} $\cdot$ \texttt{metadata\_qa}}\\
\midrule
\texttt{t2\_slices} & \ttfamily How many CT image slices are in the \{patient\_id\} study? Query the series metadata and count the instances in the CT series. Answer with only the integer count. \\
\addlinespace
\texttt{t2\_nseries} & \ttfamily How many series are in the \{patient\_id\} study (StudyInstanceUID: \{study\_uid\})? Count all series regardless of modality. Answer with only the integer count. \\
\addlinespace
\texttt{t2\_modalities} & \ttfamily What distinct imaging modalities are present in the \{patient\_id\} study? Query the series metadata and list all unique modality values, sorted alphabetically and separated by commas. Answer with only the comma-separated list (e.g., `CT, SEG, SR'), no other text. \\
\addlinespace
\texttt{t2\_date} & \ttfamily What is the study date (StudyDate DICOM tag) for the \{patient\_id\} study? Answer with only the 8-digit date in YYYYMMDD format (e.g., `20000101'), no other text. \\
\addlinespace
\texttt{t2\_ct\_uid} & \ttfamily What is the SeriesInstanceUID of the CT series in the \{patient\_id\} study? Query the series metadata and find the series with modality CT. Answer with only the SeriesInstanceUID string, no label or other text. \\
\addlinespace
\texttt{t4\_interval} & \ttfamily What is the time interval (in days) between the baseline study (StudyInstanceUID: \{baseline.study\_uid\}) and the follow-up study (StudyInstanceUID: \{followup.study\_uid\}) for participant \{participant\_id\}? Query the study metadata for both studies and compute the difference in StudyDate values. Answer with only the integer number of days, no other text. \\
\addlinespace
\texttt{t4\_slice\_diff} & \ttfamily How many more (or fewer) CT slices does the follow-up study have compared to the baseline for participant \{participant\_id\}? Baseline StudyInstanceUID: \{baseline.study\_uid\}. Follow-up StudyInstanceUID: \{followup.study\_uid\}. Query series metadata for both studies, find the CT series in each, and report the difference (follow-up minus baseline). Use a positive number if follow-up has more slices, negative if fewer. Answer with only the signed integer (e.g., `-17' or `24'), no other text. \\
\midrule
\multicolumn{2}{@{}l}{\textbf{Easy} $\cdot$ \texttt{vision\_probe}}\\
\midrule
\texttt{vp\_mod} & \ttfamily What modality is this image?\textbackslash nA) CT\textbackslash nB) MRI\textbackslash nC) DX\textbackslash nD) N/A\textbackslash n\textbackslash nRespond by calling submit\_answer with only the letter. \\
\addlinespace
\texttt{vp\_pre} & \ttfamily What windowing preset was applied to this image?\textbackslash nA) Lung window\textbackslash nB) Soft tissue window\textbackslash nC) Default\textbackslash nD) Breast MRI\textbackslash nE) N/A\textbackslash n\textbackslash nRespond by calling submit\_answer with only the letter. \\
\midrule
\multicolumn{2}{@{}l}{\textbf{Medium} $\cdot$ \texttt{annotation}}\\
\midrule
\texttt{t3\_nodule} & \ttfamily Navigate to slice \{slice\_index\} of the CT series and place a segmentation annotation on the pulmonary nodule (``\{segment\_label\}'') in this \{patient\_id\} chest CT. Apply a lung window (WW: 1500, WC: -600) for optimal visualization. Use a circle or polygon region to outline the nodule. \\
\addlinespace
\texttt{t3\_find} & \ttfamily Find and segment the nodule labeled ``\{label\}'' in this \{patient\_id\} chest CT. The nodule is visible on slices \{first\_slice\} through \{last\_slice\} (\{num\_slices\} slices). Navigate to each slice, apply a lung window, inspect the image, and place a segmentation annotation on every slice where this specific nodule is present. \\
\midrule
\multicolumn{2}{@{}l}{\textbf{Medium} $\cdot$ \texttt{oracle\_annotation}}\\
\midrule
\texttt{t3\_oracle} (single) & \ttfamily Use the external pathology detection model to identify and segment the pulmonary nodule in this \{patient\_id\} chest CT. First query the model for an overview of findings in the CT series, then request the precise segmentation contour for the recommended slice. Navigate to that slice and place the annotation using the model's output. \\
\addlinespace
\texttt{t3\_oracle} (multi) & \ttfamily Use the external pathology detection model to identify and segment all pulmonary nodules in this \{patient\_id\} chest CT. Query the model for an overview of all findings, then for each finding request the precise contour, navigate to the slice, and place an annotation. Annotate all findings. \\
\addlinespace
\texttt{t3\_oracle} (volumetric) & \ttfamily Use the external pathology detection model to segment the nodule ``\{segment\_label\}'' across all its slices in this \{patient\_id\} chest CT. Query the model for an overview, then for each slice in the nodule's range, request the precise contour, navigate to the slice, and place the annotation. \\
\midrule
\multicolumn{2}{@{}l}{\textbf{Medium} $\cdot$ \texttt{oracle\_birads\_report}}\\
\midrule
\texttt{t3\_oracle\_birads} & \ttfamily You are viewing a breast MRI study for patient \{patient\_id\}. An external breast MRI CAD model is available via the query\_birads\_model tool. Query the model for the loaded series to obtain BI-RADS findings, then submit a structured BI-RADS report using submit\_birads\_report with the model's findings including laterality, lesion count, BI-RADS category, and whether enhancement is present. \\
\midrule
\multicolumn{2}{@{}l}{\textbf{Hard} $\cdot$ \texttt{longitudinal}}\\
\midrule
\texttt{t4\_lesion} (single) & \ttfamily Compare baseline and follow-up chest CTs for participant \{participant\_id\}. A new lesion has appeared on the follow-up scan that was not present on the baseline. Baseline StudyInstanceUID: \{baseline.study\_uid\} (series: \{baseline.series\_uid\}). Follow-up StudyInstanceUID: \{followup.study\_uid\} (series: \{followup.series\_uid\}). First examine the baseline CT to understand the normal anatomy, then switch to the follow-up CT and navigate through slices to find the new lesion. Use get\_dicom\_image with lung\_window preprocessor to view the images. Once you locate the new lesion, submit its location using submit\_longitudinal\_finding. \\
\addlinespace
\texttt{t4\_lesion} (multi) & \ttfamily Compare baseline and follow-up chest CTs for participant \{participant\_id\}. Multiple new lesions may have appeared on the follow-up scan. Baseline StudyInstanceUID: \{baseline.study\_uid\} (series: \{baseline.series\_uid\}). Follow-up StudyInstanceUID: \{followup.study\_uid\} (series: \{followup.series\_uid\}). Examine both studies using lung window settings, identify all new findings, and submit each one using submit\_longitudinal\_finding. When done, call submit\_longitudinal\_complete. \\
\midrule
\multicolumn{2}{@{}l}{\textbf{Hard} $\cdot$ \texttt{birads\_report}}\\
\midrule
\texttt{t4\_birads} & \ttfamily You are viewing a breast MRI study (StudyInstanceUID: \{study\_uid\}). This study contains \{n\_series\} MR series including dynamic contrast-enhanced (DCE) sequences.\textbackslash n\textbackslash nAvailable MR series:\textbackslash n\{series\_list\}\textbackslash n\textbackslash nNavigate through the available series to identify any enhancing lesions. Compare pre-contrast and post-contrast sequences to assess enhancement. Produce a structured BI-RADS report by calling submit\_birads\_report with your findings including laterality, lesion count, BI-RADS category, and whether enhancement is present. \\

\end{longtable}
\end{small}

\paragraph{Per-task tool availability.} Each task type fixes the tool set passed to the agent for the duration of an episode. Table \ref{tab:tools-by-task-type} lists the full mapping. Tools that are irrelevant to a task's expected trajectory (for example, a reporting action in a segmentation task) are included whenever the task type admits them, so that planning can be penalised for inappropriate tool selection. Vision-probe episodes expose only a single submission tool, reflecting their classification-only nature; oracle variants substitute an oracle-query tool for the annotation primitives; and only longitudinal tasks expose the two-step finding/complete reporting pair.

\begin{table}[ht]
\centering
\footnotesize
\setlength{\tabcolsep}{3pt}
\caption{Tool availability by task type in ABRA. A \checkmark{} indicates that the tool is included in the tool set passed to the agent when a task of that type is loaded. Columns are ordered by difficulty tier (Easy, Medium, Hard).}
\label{tab:tools-by-task-type}
\resizebox{\textwidth}{!}{%
\begin{tabular}{@{}l cccccccc@{}}
\toprule
& \rotatebox[origin=lB]{60}{\texttt{viewer\_control}}
& \rotatebox[origin=lB]{60}{\texttt{metadata\_qa}}
& \rotatebox[origin=lB]{60}{\texttt{vision\_probe}}
& \rotatebox[origin=lB]{60}{\texttt{annotation}}
& \rotatebox[origin=lB]{60}{\texttt{oracle\_annotation}}
& \rotatebox[origin=lB]{60}{\texttt{oracle\_birads\_report}}
& \rotatebox[origin=lB]{60}{\texttt{longitudinal}}
& \rotatebox[origin=lB]{60}{\texttt{birads\_report}} \\[3.2em]
\midrule
\multicolumn{9}{@{}l}{\emph{Navigation and display}}\\
\texttt{set\_viewport\_slice}         & \checkmark & \checkmark &            & \checkmark & \checkmark & \checkmark & \checkmark & \checkmark \\
\texttt{set\_window\_level}           & \checkmark & \checkmark &            & \checkmark & \checkmark & \checkmark & \checkmark & \checkmark \\
\texttt{set\_zoom}                    & \checkmark & \checkmark &            & \checkmark & \checkmark & \checkmark & \checkmark & \checkmark \\
\texttt{select\_series}               & \checkmark & \checkmark &            & \checkmark & \checkmark & \checkmark & \checkmark & \checkmark \\
\midrule
\multicolumn{9}{@{}l}{\emph{Metadata queries}}\\
\texttt{get\_viewport\_state}         & \checkmark & \checkmark &            & \checkmark & \checkmark & \checkmark & \checkmark & \checkmark \\
\texttt{get\_study\_metadata}         &            & \checkmark &            & \checkmark & \checkmark & \checkmark & \checkmark & \checkmark \\
\texttt{get\_study\_series}           &            & \checkmark &            & \checkmark & \checkmark & \checkmark & \checkmark & \checkmark \\
\texttt{get\_series\_metadata}        &            & \checkmark &            & \checkmark & \checkmark & \checkmark & \checkmark & \checkmark \\
\texttt{get\_instance\_metadata}      &            & \checkmark &            & \checkmark & \checkmark & \checkmark & \checkmark & \checkmark \\
\midrule
\multicolumn{9}{@{}l}{\emph{Pixel and viewer snapshot}}\\
\texttt{get\_dicom\_image}            &            &            &            & \checkmark &            &            & \checkmark & \checkmark \\
\texttt{get\_viewer\_screenshot}      &            &            &            & \checkmark &            &            & \checkmark & \checkmark \\
\midrule
\multicolumn{9}{@{}l}{\emph{Oracle predictions}}\\
\texttt{query\_pathology\_model}      &            &            &            &            & \checkmark &            &            &            \\
\texttt{query\_birads\_model}         &            &            &            &            &            & \checkmark &            &            \\
\midrule
\multicolumn{9}{@{}l}{\emph{Segmentation}}\\
\texttt{add\_circle\_segmentation}    &            &            &            & \checkmark &            &            & \checkmark & \checkmark \\
\texttt{add\_rectangle\_segmentation} &            &            &            & \checkmark &            &            & \checkmark & \checkmark \\
\texttt{add\_polygon\_segmentation}   &            &            &            & \checkmark & \checkmark &            & \checkmark & \checkmark \\
\texttt{list\_segmentations}          &            &            &            & \checkmark & \checkmark &            & \checkmark & \checkmark \\
\midrule
\multicolumn{9}{@{}l}{\emph{Reporting}}\\
\texttt{submit\_answer}               &            & \checkmark & \checkmark & \checkmark & \checkmark & \checkmark & \checkmark & \checkmark \\
\texttt{submit\_birads\_report}       &            &            &            &            &            & \checkmark &            & \checkmark \\
\texttt{submit\_longitudinal\_finding}&            &            &            &            &            &            & \checkmark &            \\
\texttt{submit\_longitudinal\_complete}&           &            &            &            &            &            & \checkmark &            \\
\bottomrule
\end{tabular}%
}
\end{table}

\subsection{Initial State Setup Details}
\label{apd:initial-state}

Before each episode the controller issues a \texttt{task\_reset} call to the viewer, which clears all prior annotations and in-memory viewport state, loads the task's \texttt{study\_uid} (together with any auxiliary studies such as baseline/follow-up pairs for longitudinal tasks), selects the initial series specified by \texttt{initial\_series\_uid}, and navigates to \texttt{initial\_slice\_index}. The controller then queries \texttt{get\_viewport\_state} and embeds the result in the system prompt so the agent starts from a concrete, deterministic snapshot of the loaded study, isolating evaluation from ordering effects or leakage across episodes.

\subsection{Evaluation}
\label{apd:evaluation-details}

Each episode is scored on three dimensions whose weighted sum defines the per-task score:
\[
S \;=\; 0.20\cdot P \;+\; 0.30\cdot E \;+\; 0.50\cdot O,\qquad S,\,P,\,E,\,O\in[0,1].
\]
All three dimensions are evaluated for every task type; only the \emph{Outcome} scorer is task-type specific.

\subsubsection{Planning}
\label{apd:planning}
Planning compares the agent's tool-call sequence to the task's reference trajectory as an \emph{unordered multiset}:
\[
P \;=\; \max\!\left(0,\;\;F_1(T_{\text{ref}},\,T_{\text{agent}}) \;-\; \min\!\bigl(0.30,\;0.05\cdot\max(0,\,|T_{\text{agent}}|-|T_{\text{ref}}|)\bigr)\right),
\]
where $F_1$ is computed over tool-name counts (precision $=\mathrm{TP}/|T_{\text{agent}}|$, recall $=\mathrm{TP}/|T_{\text{ref}}|$, $\mathrm{TP}=\sum_{t}\min(\mathrm{ref}[t],\mathrm{act}[t])$). The subtracted term is a \emph{redundancy penalty} of $0.05$ for each tool call beyond the reference length, capped at $0.30$. Ordering is intentionally ignored so that agents can interleave exploratory reads without being penalised on Planning.

Reference trajectories $T_{\text{ref}}$ are constructed deterministically by each task generator from the minimum tool sequence needed to satisfy the task; they are not learned from agent rollouts. Table~\ref{tab:reference-trajectories} lists the construction for every task subtype. Some trajectories have a length that depends on per-task parameters: $N$ is the number of slices a nodule spans, $K$ is the number of findings returned by the oracle overview, $L$ is the number of ground-truth lesions in a longitudinal pair, and $V$ is the number of MR series shown to the agent (capped at $4$).

\begin{table}[ht]
\centering
\footnotesize
\setlength{\tabcolsep}{3pt}
\caption{Reference trajectories used by the planning scorer, by task subtype. Tool names are compared as an \emph{unordered multiset}, so ordering inside a row is shown only for readability. Bracketed blocks with a superscript (for example $[\,\text{svs},\text{gdi}\,]^{5}$) are repeated inline.}
\label{tab:reference-trajectories}
\resizebox{\textwidth}{!}{%
\begin{tabular}{@{}l p{11cm} l@{}}
\toprule
\textbf{Task ID prefix} & \textbf{Reference trajectory} & \textbf{Length} \\
\midrule
\multicolumn{3}{@{}l}{\textbf{Easy}}\\
\texttt{t1\_slice}        & {\ttfamily set\_viewport\_slice}                                                                                            & 1 \\
\texttt{t1\_wl\_*}        & {\ttfamily set\_window\_level}                                                                                              & 1 \\
\texttt{t1\_slice\_wl}    & {\ttfamily set\_viewport\_slice, set\_window\_level}                                                                        & 2 \\
\texttt{t1\_series}       & {\ttfamily get\_study\_series, select\_series}                                                                              & 2 \\
\texttt{t2\_slices}, \texttt{t2\_modalities}, \texttt{t2\_ct\_uid} & {\ttfamily get\_study\_series, submit\_answer}                                             & 2 \\
\texttt{t2\_nseries}, \texttt{t2\_date} & {\ttfamily get\_study\_metadata, submit\_answer}                                                              & 2 \\
\texttt{t4\_interval}     & {\ttfamily get\_study\_metadata, get\_study\_metadata, submit\_answer}                                                      & 3 \\
\texttt{t4\_slice\_diff}  & {\ttfamily get\_study\_series, get\_study\_series, submit\_answer}                                                          & 3 \\
\texttt{vp\_mod}, \texttt{vp\_pre} & {\ttfamily submit\_answer}                                                                                         & 1 \\
\midrule
\multicolumn{3}{@{}l}{\textbf{Medium}}\\
\texttt{t3\_nodule}       & {\ttfamily get\_study\_series, set\_viewport\_slice, set\_window\_level, get\_dicom\_image, add\_circle\_segmentation}       & 5 \\
\texttt{t3\_find}         & {\ttfamily get\_study\_series, set\_viewport\_slice, set\_window\_level, get\_dicom\_image, add\_circle\_segmentation,} $[\,${\ttfamily set\_viewport\_slice, get\_dicom\_image, add\_circle\_segmentation}$\,]^{\,N-1}$ & $5 + 3(N{-}1)$ \\
\texttt{t3\_oracle} (single) & {\ttfamily get\_study\_series, query\_pathology\_model, query\_pathology\_model, set\_viewport\_slice, add\_polygon\_segmentation} & 5 \\
\texttt{t3\_oracle} (multi)  & {\ttfamily get\_study\_series, query\_pathology\_model,} $[\,${\ttfamily query\_pathology\_model, set\_viewport\_slice, add\_polygon\_segmentation}$\,]^{\,K}$ & $2 + 3K$ \\
\texttt{t3\_oracle} (volumetric) & {\ttfamily get\_study\_series, query\_pathology\_model,} $[\,${\ttfamily query\_pathology\_model, set\_viewport\_slice, add\_polygon\_segmentation}$\,]^{\,N}$ & $2 + 3N$ \\
\texttt{t3\_oracle\_birads} & {\ttfamily get\_study\_series, query\_birads\_model, submit\_birads\_report}                                                & 3 \\
\midrule
\multicolumn{3}{@{}l}{\textbf{Hard}}\\
\texttt{t4\_lesion} (single) & {\ttfamily get\_study\_series, select\_series, set\_window\_level,} $[\,${\ttfamily set\_viewport\_slice, get\_dicom\_image}$\,]^{\,5}${\ttfamily , select\_series, set\_viewport\_slice, set\_window\_level,} $[\,${\ttfamily set\_viewport\_slice, get\_dicom\_image}$\,]^{\,5}${\ttfamily , submit\_longitudinal\_finding, submit\_longitudinal\_complete} & 28 \\
\texttt{t4\_lesion} (multi)  & {\ttfamily get\_study\_series, select\_series, set\_window\_level,} $[\,${\ttfamily set\_viewport\_slice, get\_dicom\_image}$\,]^{\,5}${\ttfamily , select\_series, set\_viewport\_slice, set\_window\_level,} $[\,${\ttfamily set\_viewport\_slice, get\_dicom\_image}$\,]^{\,5}${\ttfamily ,} $[\,${\ttfamily submit\_longitudinal\_finding}$\,]^{\,L}${\ttfamily , submit\_longitudinal\_complete} & $27 + L$ \\
\texttt{t4\_birads}       & {\ttfamily get\_study\_series,} $\bigl[\,${\ttfamily select\_series,} $[\,${\ttfamily set\_viewport\_slice, get\_dicom\_image}$\,]^{\,3}\bigr]^{\,V}${\ttfamily , submit\_birads\_report} & $2 + 7V$ \\
\bottomrule
\end{tabular}%
}
\end{table}

\subsubsection{Execution}
\label{apd:execution}
Execution aggregates four facets of how well the chosen tools were run (weights in parentheses):
\[
E \;=\; 0.40\cdot A_{\text{tool}} \;+\; 0.20\cdot Q_{\text{param}} \;+\; 0.25\cdot E_{\text{turn}} \;+\; 0.15\cdot R_{\text{err}}.
\]
\begin{itemize}
\setlength\itemsep{1pt}
  \item $A_{\text{tool}}$: tool-call accuracy, the fraction of calls that returned success.
  \item $Q_{\text{param}}$: parameter quality, a task-aware rubric checking that arguments are well-typed and consistent with the loaded study (for example, a window/level pair plausible for the modality or a slice index inside $[0,\,\text{totalImages})$).
  \item $E_{\text{turn}} = \min\bigl(1,\;|T_{\text{ref}}|/\max(1,\,n_{\text{turns}})\bigr)$: turn efficiency relative to the reference length.
  \item $R_{\text{err}}$: error-recovery score, rewarding agents that retry or replan after a failed tool call rather than repeating it identically.
\end{itemize}

\subsubsection{Outcome}
\label{apd:outcome}
The outcome scorer is selected per task type (Table~\ref{tab:outcome-scorers-full}). All outcome scores are clipped to $[0,1]$.

\begin{table}[ht]
\centering
\small
\caption{Task-type-specific outcome scorers used in ABRA.}
\label{tab:outcome-scorers-full}
\setlength{\tabcolsep}{3pt}
\begin{tabular}{@{}l p{3.6cm} p{7.4cm}@{}}
\toprule
\textbf{Scorer} & \textbf{Task types} & \textbf{Signal} \\
\midrule
State diff          & \texttt{viewer\_control} & Per-field match against \texttt{expected\_outcome} on the final viewport state; partial credit for multi-field tasks. \\
Exact match         & \texttt{metadata\_qa}, \texttt{vision\_probe} & Case- and whitespace-normalised equality between the submitted answer and the reference string. \\
IoU (normalised)    & \texttt{annotation}, \texttt{oracle\_annotation} & Raw intersection-over-union against the reference mask, divided by the IoU of the best-fit primitive of the agent's chosen shape: an area-matched circle centred on the reference centroid for circles; the tighter of the axis-aligned and minimum rotated bounding box for rectangles; fixed at $1$ for polygons. A separate hit rate at normalised IoU$\geq 0.5$ is also reported. \\
Point distance      & \texttt{longitudinal} (single finding) & $1$ within a $20$-pixel radius of the reference point, linear falloff to $0$ at $40$\,px, hard zero if the submission names the wrong slice. \\
Multi-finding       & \texttt{longitudinal} (multiple findings) & Detection rate minus $0.1$ per false positive, clamped to $[0,1]$; matching uses the same slice and distance criterion as above. \\
BI-RADS report      & \texttt{birads\_report}, \texttt{oracle\_birads\_report} & Weighted field comparison (Table~\ref{tab:birads-fields}). \\
\bottomrule
\end{tabular}
\end{table}

\paragraph{BI-RADS field weights.} The BI-RADS report scorer is a weighted average over five fields. The assessment category and lesion count allow partial credit on off-by-one errors ($0.5$ for $|\Delta|=1$), reflecting the ordinal nature of the BI-RADS scale and the clinical tolerability of small count discrepancies.

\begin{table}[ht]
\centering
\small
\caption{Scored fields in the BI-RADS report scorer. Unscored free-text fields (per-lesion shape, margin, enhancement, type, size; textual recommendation) are retained in the trace for qualitative analysis.}
\label{tab:birads-fields}
\setlength{\tabcolsep}{6pt}
\begin{tabular}{@{}l r l@{}}
\toprule
\textbf{Field} & \textbf{Weight} & \textbf{Match rule} \\
\midrule
\texttt{laterality}         & 0.25 & Exact match (left/right/bilateral). \\
\texttt{birads\_category}   & 0.30 & Exact, or $0.5$ for $|\Delta|=1$. \\
\texttt{lesion\_count}      & 0.20 & Exact, or $0.5$ for $|\Delta|=1$. \\
\texttt{enhancement\_present} & 0.15 & Boolean match. \\
\texttt{lesion\_quadrant}   & 0.10 & Exact match; omitted from the denominator when the reference is absent. \\
\bottomrule
\end{tabular}
\end{table}

\section{Baseline methods details}
\label{apd:baselines}

\subsection{Per-task-type score breakdown}
\label{apd:per-task-results}

Table~\ref{tab:per-task-results} gives the per-task-type Planning, Execution, and Outcome scores underlying the tier averages in Table~\ref{tab:abra-results}. \texttt{vision\_probe} is the only Easy task whose Outcome stays below 1.0 on every model, since modality and preprocessor identification is itself a small visual classification rather than a tag lookup or a viewport command. On the Hard tier, achieved Outcome concentrates on \texttt{birads\_report} (0.06--0.64 across the roster) and is uniformly below 0.03 on \texttt{longitudinal}.

\begin{longtable}{@{}l l r r r r r@{}}
\caption{Per-model, per-task-type Planning, Execution, and Outcome scores across the eight ABRA task types, grouped by difficulty tier. $n$ is the number of tasks of the type.}
\label{tab:per-task-results} \\
\toprule
\textbf{Task type} & \textbf{Model} & $n$ & \textbf{Plan $\uparrow$} & \textbf{Exec $\uparrow$} & \textbf{Outcome $\uparrow$} & \textbf{Avg $\uparrow$} \\
\midrule
\endfirsthead
\multicolumn{7}{l}{\textit{Table~\ref{tab:per-task-results} continued from previous page}} \\
\toprule
\textbf{Task type} & \textbf{Model} & $n$ & \textbf{Plan $\uparrow$} & \textbf{Exec $\uparrow$} & \textbf{Outcome $\uparrow$} & \textbf{Avg $\uparrow$} \\
\midrule
\endhead
\midrule
\multicolumn{7}{r}{\textit{continued on next page}} \\
\endfoot
\bottomrule
\endlastfoot
\multicolumn{7}{@{}l}{\textbf{Easy}}\\
\midrule
\texttt{viewer\_control}        & Claude Sonnet 4.6        & 60  & 0.98 & 1.00 & 1.00 & 1.00 \\
                                & GPT-5.4                  & 60  & 0.87 & 0.98 & 1.00 & 0.97 \\
                                & GPT-5.4-nano             & 60  & 0.94 & 0.98 & 1.00 & 0.98 \\
                                & Gemini 3 Flash  & 60  & 0.11 & 0.77 & 0.95 & 0.73 \\
                                & Gemini 3 Pro    & 60  & 0.02 & 0.78 & 0.98 & 0.73 \\
                                & Qwen 3.5                 & 60  & 0.93 & 0.98 & 1.00 & 0.98 \\
                                & Gemma 4                  & 60  & 0.99 & 1.00 & 1.00 & 1.00 \\
                                & Ministral 3 (14B)        & 60  & 0.94 & 0.99 & 1.00 & 0.99 \\
                                & Mistral Large 3          & 60  & 0.75 & 0.92 & 0.97 & 0.91 \\
                                & Kimi K2.5                & 60  & 0.20 & 0.83 & 1.00 & 0.79 \\
\addlinespace
\texttt{metadata\_qa}           & Claude Sonnet 4.6        & 90  & 0.82 & 1.00 & 1.00 & 0.96 \\
                                & GPT-5.4                  & 90  & 0.64 & 1.00 & 0.87 & 0.86 \\
                                & GPT-5.4-nano             & 90  & 0.90 & 1.00 & 0.98 & 0.97 \\
                                & Gemini 3 Flash  & 90  & 0.51 & 0.96 & 0.98 & 0.88 \\
                                & Gemini 3 Pro    & 90  & 0.59 & 0.99 & 0.82 & 0.83 \\
                                & Qwen 3.5                 & 90  & 0.71 & 1.00 & 1.00 & 0.94 \\
                                & Gemma 4                  & 90  & 0.69 & 1.00 & 0.78 & 0.83 \\
                                & Ministral 3 (14B)        & 90  & 0.82 & 0.98 & 0.56 & 0.74 \\
                                & Mistral Large 3          & 90  & 0.89 & 1.00 & 0.79 & 0.87 \\
                                & Kimi K2.5                & 90  & 0.96 & 1.00 & 0.87 & 0.93 \\
\addlinespace
\texttt{vision\_probe}          & Claude Sonnet 4.6        & 99  & 0.99 & 0.99 & 0.65 & 0.82 \\
                                & GPT-5.4                  & 99  & 0.99 & 0.99 & 0.83 & 0.91 \\
                                & GPT-5.4-nano             & 99  & 0.99 & 0.99 & 0.70 & 0.84 \\
                                & Gemini 3 Flash  & 99  & 1.00 & 1.00 & 0.75 & 0.87 \\
                                & Gemini 3 Pro    & 99  & 1.00 & 1.00 & 0.64 & 0.82 \\
                                & Qwen 3.5                 & 99  & 1.00 & 1.00 & 0.77 & 0.88 \\
                                & Gemma 4                  & 99  & 1.00 & 1.00 & 0.88 & 0.94 \\
                                & Ministral 3 (14B)        & 99  & 1.00 & 1.00 & 0.61 & 0.80 \\
                                & Mistral Large 3          & 99  & 1.00 & 1.00 & 0.43 & 0.72 \\
                                & Kimi K2.5                & 99  & 0.83 & 0.83 & 0.48 & 0.66 \\
\midrule
\multicolumn{7}{@{}l}{\textbf{Medium}}\\
\midrule
\texttt{annotation}             & Claude Sonnet 4.6        & 151 & 0.73 & 0.99 & 0.02 & 0.45 \\
                                & GPT-5.4                  & 151 & 0.80 & 0.99 & 0.03 & 0.47 \\
                                & GPT-5.4-nano             & 151 & 0.66 & 0.98 & 0.00 & 0.42 \\
                                & Gemini 3 Flash  & 151 & 0.57 & 0.96 & 0.25 & 0.53 \\
                                & Gemini 3 Pro    & 151 & 0.65 & 0.99 & 0.16 & 0.51 \\
                                & Qwen 3.5                 & 151 & 0.73 & 0.89 & 0.00 & 0.42 \\
                                & Gemma 4                  & 151 & 0.74 & 0.91 & 0.08 & 0.46 \\
                                & Ministral 3 (14B)        & 151 & 0.74 & 0.99 & 0.00 & 0.45 \\
                                & Mistral Large 3          & 151 & 0.77 & 0.98 & 0.00 & 0.45 \\
                                & Kimi K2.5                & 151 & 0.70 & 0.99 & 0.02 & 0.45 \\
\addlinespace
\texttt{oracle\_annotation}     & Claude Sonnet 4.6        & 51  & 0.90 & 1.00 & 1.00 & 0.98 \\
                                & GPT-5.4                  & 51  & 0.90 & 1.00 & 0.98 & 0.97 \\
                                & GPT-5.4-nano             & 51  & 0.80 & 1.00 & 0.96 & 0.94 \\
                                & Gemini 3 Flash  & 51  & 0.53 & 0.88 & 0.90 & 0.82 \\
                                & Gemini 3 Pro    & 51  & 0.53 & 0.94 & 0.69 & 0.73 \\
                                & Qwen 3.5                 & 51  & 0.82 & 1.00 & 0.95 & 0.94 \\
                                & Gemma 4                  & 51  & 0.72 & 0.95 & 0.88 & 0.87 \\
                                & Ministral 3 (14B)        & 51  & 0.87 & 0.96 & 0.90 & 0.91 \\
                                & Mistral Large 3          & 51  & 0.89 & 1.00 & 0.91 & 0.93 \\
                                & Kimi K2.5                & 51  & 0.51 & 0.93 & 0.84 & 0.80 \\
\addlinespace
\texttt{oracle\_birads\_report} & Claude Sonnet 4.6        & 50  & 0.80 & 1.00 & 1.00 & 0.96 \\
                                & GPT-5.4                  & 50  & 0.80 & 1.00 & 1.00 & 0.96 \\
                                & GPT-5.4-nano             & 50  & 0.80 & 1.00 & 1.00 & 0.96 \\
                                & Gemini 3 Flash  & 50  & 0.67 & 1.00 & 1.00 & 0.93 \\
                                & Gemini 3 Pro    & 50  & 0.61 & 1.00 & 0.58 & 0.71 \\
                                & Qwen 3.5                 & 50  & 0.80 & 1.00 & 1.00 & 0.96 \\
                                & Gemma 4                  & 50  & 0.80 & 1.00 & 1.00 & 0.96 \\
                                & Ministral 3 (14B)        & 50  & 0.80 & 1.00 & 1.00 & 0.96 \\
                                & Mistral Large 3          & 50  & 0.80 & 1.00 & 1.00 & 0.96 \\
                                & Kimi K2.5                & 50  & 0.80 & 1.00 & 0.94 & 0.93 \\
\midrule
\multicolumn{7}{@{}l}{\textbf{Hard}}\\
\midrule
\texttt{longitudinal}           & Claude Sonnet 4.6        & 104 & 0.05 & 0.98 & 0.00 & 0.30 \\
                                & GPT-5.4                  & 104 & 0.20 & 0.98 & 0.00 & 0.34 \\
                                & GPT-5.4-nano             & 104 & 0.61 & 0.97 & 0.00 & 0.41 \\
                                & Gemini 3 Flash  & 104 & 0.42 & 0.85 & 0.00 & 0.34 \\
                                & Gemini 3 Pro    & 104 & 0.26 & 0.94 & 0.00 & 0.33 \\
                                & Qwen 3.5                 & 104 & 0.45 & 0.99 & 0.00 & 0.39 \\
                                & Gemma 4                  & 104 & 0.45 & 0.98 & 0.00 & 0.38 \\
                                & Ministral 3 (14B)        & 104 & 0.47 & 0.95 & 0.02 & 0.39 \\
                                & Mistral Large 3          & 104 & 0.55 & 0.88 & 0.00 & 0.37 \\
                                & Kimi K2.5                & 104 & 0.38 & 0.98 & 0.01 & 0.38 \\
\addlinespace
\texttt{birads\_report}         & Claude Sonnet 4.6        & 50  & 0.53 & 0.99 & 0.64 & 0.73 \\
                                & GPT-5.4                  & 50  & 0.53 & 0.87 & 0.32 & 0.53 \\
                                & GPT-5.4-nano             & 50  & 0.45 & 0.98 & 0.12 & 0.45 \\
                                & Gemini 3 Flash  & 50  & 0.27 & 0.90 & 0.18 & 0.41 \\
                                & Gemini 3 Pro    & 50  & 0.48 & 1.00 & 0.50 & 0.65 \\
                                & Qwen 3.5                 & 50  & 0.59 & 1.00 & 0.35 & 0.59 \\
                                & Gemma 4                  & 50  & 0.25 & 0.82 & 0.06 & 0.33 \\
                                & Ministral 3 (14B)        & 50  & 0.36 & 0.71 & 0.35 & 0.46 \\
                                & Mistral Large 3          & 50  & 0.43 & 0.94 & 0.47 & 0.60 \\
                                & Kimi K2.5                & 50  & 0.58 & 0.99 & 0.42 & 0.62 \\
\end{longtable}

\subsection{Segmentation task analysis}
\label{apd:segmentation-analysis}

{\footnotesize
\setlength{\tabcolsep}{3pt}
\begin{longtable}{@{}l l r r r r r r r r@{}}
\caption{Per-model annotation behaviour on LIDC single-slice and volumetric annotation variants. \emph{Slice match \%} is the fraction of tasks where at least one agent annotation lands on a ground-truth axial slice. \emph{Centroid} distances are computed only on slice-matched annotations. \emph{Volume ratio} is per-pair agent-drawn area divided by reference polygon area, summed across slices. \emph{Mode radius} is the single circle radius used most often, with the share of all circle annotations using it. Reference lesion area: median 120 px$^2$ (single-slice) and 62 px$^2$ (volumetric).}
\label{tab:annotation-behaviour} \\
\toprule
\multirow{2}{*}{\textbf{Model}} & \multirow{2}{*}{\textbf{Variant}} & \multirow{2}{*}{$n$} & \multirow{2}{*}{\textbf{Slice match \% $\uparrow$}} & \multicolumn{2}{c}{\textbf{Centroid (px) $\downarrow$}} & \multicolumn{2}{c}{\textbf{Vol.\ ratio $\downarrow$}} & \multirow{2}{*}{\textbf{Circle \%}} & \multirow{2}{*}{\textbf{Mode radius (px, \%)}} \\
\cmidrule(lr){5-6} \cmidrule(lr){7-8}
 & & & & med & p90 & med & p90 & & \\
\midrule
\endfirsthead
\multicolumn{10}{l}{\textit{Table~\ref{tab:annotation-behaviour} continued from previous page}} \\
\toprule
\multirow{2}{*}{\textbf{Model}} & \multirow{2}{*}{\textbf{Variant}} & \multirow{2}{*}{$n$} & \multirow{2}{*}{\textbf{Slice match \% $\uparrow$}} & \multicolumn{2}{c}{\textbf{Centroid (px) $\downarrow$}} & \multicolumn{2}{c}{\textbf{Vol.\ ratio $\downarrow$}} & \multirow{2}{*}{\textbf{Circle \%}} & \multirow{2}{*}{\textbf{Mode radius (px, \%)}} \\
\cmidrule(lr){5-6} \cmidrule(lr){7-8}
 & & & & med & p90 & med & p90 & & \\
\midrule
\endhead
\midrule
\multicolumn{10}{r}{\textit{continued on next page}} \\
\endfoot
\bottomrule
\endlastfoot
Claude Sonnet 4.6        & single-slice & 129 & 100  & 116 & 200 & 4.0 & 29.2 & 100  & 12 (67) \\
                         & volumetric   &  22 & 100  &  89 & 203 & 6.2 & 33.1 & 100  & 12 (25) \\
\addlinespace
GPT-5.4                  & single-slice & 129 & 98.4 &  90 & 262 & 2.1 & 14.9 & 97.6 & 8 (51) \\
                         & volumetric   &  22 & 100  &  65 & 181 & 3.3 & 17.7 & 78.2 & 7 (22) \\
\addlinespace
GPT-5.4-nano             & single-slice & 129 & 100  & 104 & 177 & 5.9 & 45.2 & 95.3 & 18 (69) \\
                         & volumetric   &  22 & 81.8 &  70 & 154 & 9.6 & 55.0 & 100  & 18 (82) \\
\addlinespace
Gemini 3 Flash  & single-slice & 129 & 98.4 &  34 & 206 & 2.7 & 16.1 & 100  & 10 (45) \\
                         & volumetric   &  22 & 100  &  55 & 188 & 2.4 & 16.1 & 100  & 5 (26) \\
\addlinespace
Gemini 3 Pro    & single-slice & 129 & 98.4 &  44 & 198 & 2.1 & 16.1 & 100  & 10 (46) \\
                         & volumetric   &  22 & 95.5 &  11 & 169 & 1.9 & 17.5 & 100  & 5 (23) \\
\addlinespace
Qwen 3.5                 & single-slice & 129 & 86.8 & 102 & 206 & 3.1 & 36.6 & 99.1 & 18 (21) \\
                         & volumetric   &  22 & 90.9 &  74 & 130 & 3.9 & 44.7 & 100  & 10 (34) \\
\addlinespace
Gemma 4                  & single-slice & 129 & 89.9 &  85 & 161 & 2.2 & 16.1 & 100  & 10 (65) \\
                         & volumetric   &  22 & 95.5 &  85 & 224 & 2.7 & 16.1 & 100  & 10 (68) \\
\addlinespace
Ministral 3 (14B)        & single-slice & 129 & 97.7 & 135 & 187 & 1.1 &  8.0 & 31.7 & 8 (55) \\
                         & volumetric   &  22 & 86.4 & 110 & 194 & 5.1 & 33.1 & 96.9 & 10 (31) \\
\addlinespace
Mistral Large 3          & single-slice & 129 & 100  & 124 & 184 & 4.2 & 45.2 & 39.5 & 12 (49) \\
                         & volumetric   &  22 & 100  & 118 & 194 & 2.6 & 16.1 & 98.9 & 8 (68) \\
\addlinespace
Kimi K2.5                & single-slice & 129 & 99.2 &  63 & 208 & 3.8 & 25.1 & 100  & 12 (39) \\
                         & volumetric   &  22 & 100  &  72 & 188 & 2.9 & 16.4 & 100  & 8 (27) \\
\end{longtable}
}

\subsection{Oracle annotation analysis}
\label{apd:oracle-analysis}

{\footnotesize
\setlength{\tabcolsep}{3pt}
\begin{longtable}{@{}l l r r r r r r r r@{}}
\caption{Per-model behaviour on LIDC oracle annotation tasks. The oracle tool returns the reference polygon (or per-slice polygons for the volumetric variant), and the task requires the agent to deposit those polygons in the viewer. \emph{Queried} is the fraction of tasks where the agent invoked the oracle tool. Each task is then assigned to one of three mutually exclusive outcome buckets: \emph{abstain} (no annotation submitted), \emph{complete} (oracle polygons submitted verbatim on the correct slices), or alter (the residual, defined as $100 - \text{abstain} - \text{complete}$). The four columns under \emph{Alter} characterise the alter cases and may overlap rather than sum: \emph{wrong slice} (single-slice only) records placement on a non-ground-truth axial slice; \emph{points modified} records submissions whose polygon differs from the oracle's; \emph{extra} records submissions of more annotations than the oracle returned; \emph{partial} (volumetric only) records cases where at least one ground-truth slice received no annotation.}
\label{tab:oracle-behaviour} \\
\toprule
\multirow{2}{*}{\textbf{Model}} & \multirow{2}{*}{\textbf{Variant}} & \multirow{2}{*}{$n$} & \multirow{2}{*}{\textbf{Queried \% $\uparrow$}} & \multirow{2}{*}{\textbf{Abstain \% $\downarrow$}} & \multirow{2}{*}{\textbf{Complete \% $\uparrow$}} & \multicolumn{4}{c}{\textbf{Alter (\% of tasks) $\downarrow$}} \\
\cmidrule(lr){7-10}
 & & & & & & \textbf{Wrong slice} & \textbf{Points modified} & \textbf{Extra} & \textbf{Partial} \\
\midrule
\endfirsthead
\multicolumn{10}{l}{\textit{Table~\ref{tab:oracle-behaviour} continued from previous page}} \\
\toprule
\multirow{2}{*}{\textbf{Model}} & \multirow{2}{*}{\textbf{Variant}} & \multirow{2}{*}{$n$} & \multirow{2}{*}{\textbf{Queried \% $\uparrow$}} & \multirow{2}{*}{\textbf{Abstain \% $\downarrow$}} & \multirow{2}{*}{\textbf{Complete \% $\uparrow$}} & \multicolumn{4}{c}{\textbf{Alter (\% of tasks) $\downarrow$}} \\
\cmidrule(lr){7-10}
 & & & & & & \textbf{Wrong slice} & \textbf{Points modified} & \textbf{Extra} & \textbf{Partial} \\
\midrule
\endhead
\midrule
\multicolumn{10}{r}{\textit{continued on next page}} \\
\endfoot
\bottomrule
\endlastfoot
Claude Sonnet 4.6        & single-slice & 23 & 100 & 0    & 100  & 0   & 0    & 0 & n/a  \\
                         & volumetric   & 21 & 100 & 0    & 100  & n/a & 0    & 0 & 0    \\
\addlinespace
GPT-5.4                  & single-slice & 23 & 100 & 0    & 100  & 0   & 0    & 0 & n/a  \\
                         & volumetric   & 21 & 100 & 0    & 90.5 & n/a & 0    & 0 & 9.5  \\
\addlinespace
GPT-5.4-nano             & single-slice & 23 & 100 & 0    & 82.6 & 0   & 17.4 & 0 & n/a  \\
                         & volumetric   & 21 & 100 & 0    & 47.6 & n/a & 23.8 & 0 & 38.1 \\
\addlinespace
Gemini 3 Flash  & single-slice & 23 & 100 & 0    & 100  & 0   & 0    & 0 & n/a  \\
                         & volumetric   & 21 & 100 & 0    & 100  & n/a & 0    & 0 & 0    \\
\addlinespace
Gemini 3 Pro    & single-slice & 23 & 100 & 13   & 87   & 0   & 0    & 0 & n/a  \\
                         & volumetric   & 21 & 100 & 52.4 & 47.6 & n/a & 0    & 0 & 52.4 \\
\addlinespace
Qwen 3.5                 & single-slice & 23 & 100 & 0    & 100  & 0   & 0    & 0 & n/a  \\
                         & volumetric   & 21 & 100 & 0    & 95.2 & n/a & 4.8  & 0 & 0    \\
\addlinespace
Gemma 4                  & single-slice & 23 & 100 & 0    & 56.5 & 0   & 43.5 & 0 & n/a  \\
                         & volumetric   & 21 & 100 & 4.8  & 66.7 & n/a & 28.6 & 0 & 9.5  \\
\addlinespace
Ministral 3 (14B)        & single-slice & 23 & 100 & 0    & 95.7 & 0   & 4.3  & 0 & n/a  \\
                         & volumetric   & 21 & 100 & 0    & 66.7 & n/a & 33.3 & 0 & 4.8  \\
\addlinespace
Mistral Large 3          & single-slice & 23 & 100 & 0    & 100  & 0   & 0    & 0 & n/a  \\
                         & volumetric   & 21 & 100 & 0    & 57.1 & n/a & 4.8  & 0 & 42.9 \\
\addlinespace
Kimi K2.5                & single-slice & 23 & 100 & 21.7 & 78.3 & 0   & 0    & 0 & n/a  \\
                         & volumetric   & 21 & 100 & 0    & 52.4 & n/a & 14.3 & 0 & 42.9 \\
\end{longtable}
}

\subsection{Oracle BI-RADS analysis}
\label{apd:oracle-birads-analysis}

{\footnotesize
\setlength{\tabcolsep}{3pt}
\begin{longtable}{@{}l r r r r r r r r r@{}}
\caption{Per-model behaviour on Duke oracle BI-RADS reporting tasks ($n=50$). The oracle tool returns the reference report fields; the task requires the agent to submit those fields verbatim. \emph{Queried} is the fraction of tasks where the agent invoked the oracle tool. Each task is then assigned to one of three mutually exclusive outcome buckets: \emph{abstain} (no report submitted), \emph{complete} (oracle fields submitted verbatim), or alter (the residual). The five columns under \emph{Alter} record which specific field of the report was modified. Surprisingly, Gemini 3 Pro always queried the oracle yet did not submit a final report in 42\% of tasks, yielding Outcome\,$=\,0$ on those episodes.}
\label{tab:oracle-birads-behaviour} \\
\toprule
\multirow{2}{*}{\textbf{Model}} & \multirow{2}{*}{$n$} & \multirow{2}{*}{\textbf{Queried \% $\uparrow$}} & \multirow{2}{*}{\textbf{Abstain \% $\downarrow$}} & \multirow{2}{*}{\textbf{Complete \% $\uparrow$}} & \multicolumn{5}{c}{\textbf{Alter (\% of tasks) $\downarrow$}} \\
\cmidrule(lr){6-10}
 & & & & & \textbf{Laterality} & \textbf{Lesion count} & \textbf{Category} & \textbf{Enhancement} & \textbf{Quadrant} \\
\midrule
\endfirsthead
\multicolumn{10}{l}{\textit{Table~\ref{tab:oracle-birads-behaviour} continued from previous page}} \\
\toprule
\multirow{2}{*}{\textbf{Model}} & \multirow{2}{*}{$n$} & \multirow{2}{*}{\textbf{Queried \% $\uparrow$}} & \multirow{2}{*}{\textbf{Abstain \% $\downarrow$}} & \multirow{2}{*}{\textbf{Complete \% $\uparrow$}} & \multicolumn{5}{c}{\textbf{Alter (\% of tasks) $\downarrow$}} \\
\cmidrule(lr){6-10}
 & & & & & \textbf{Laterality} & \textbf{Lesion count} & \textbf{Category} & \textbf{Enhancement} & \textbf{Quadrant} \\
\midrule
\endhead
\midrule
\multicolumn{10}{r}{\textit{continued on next page}} \\
\endfoot
\bottomrule
\endlastfoot
Claude Sonnet 4.6 & 50 & 100 & 0  & 100 & 0 & 0 & 0 & 0 & 0  \\
GPT-5.4           & 50 & 100 & 0  & 100 & 0 & 0 & 0 & 0 & 0  \\
GPT-5.4-nano      & 50 & 100 & 0  & 100 & 0 & 0 & 0 & 0 & 0  \\
Gemini 3 Flash    & 50 & 100 & 0  & 100 & 0 & 0 & 0 & 0 & 0  \\
Gemini 3 Pro      & 50 & 100 & 42 & 58  & 0 & 0 & 0 & 0 & 0  \\
Qwen 3.5          & 50 & 100 & 0  & 100 & 0 & 0 & 0 & 0 & 0  \\
Gemma 4           & 50 & 100 & 0  & 98  & 0 & 0 & 0 & 0 & 2  \\
Ministral 3 (14B) & 50 & 100 & 0  & 100 & 0 & 0 & 0 & 0 & 0  \\
Mistral Large 3   & 50 & 100 & 0  & 100 & 0 & 0 & 0 & 0 & 0  \\
Kimi K2.5         & 50 & 100 & 0  & 38  & 0 & 0 & 0 & 0 & 62 \\
\end{longtable}
}

\subsection{Longitudinal task analysis}
\label{apd:longitudinal-analysis}

{\footnotesize
\setlength{\tabcolsep}{3pt}
\begin{longtable}{@{}l r r r r r r r r r@{}}
\caption{Per-model localisation error on single new-lesion longitudinal tasks ($n=90$, NLST baseline-and-follow-up CT pairs). The agent must identify the new lesion appearing on the follow-up study and submit a single in-plane point on the slice carrying the finding. \emph{Submitted} is the fraction of tasks with any submission. \emph{Correct slice} requires exact match between submitted and ground-truth slice indices. \emph{Slice diff}, \emph{3D mm}, and \emph{Within 100 mm} are computed over the submitted subset only. \emph{3D mm} is the Euclidean distance combining in-plane error (PixelSpacing) and axial error (slice spacing) from per-series DICOM headers. \emph{Hit} is the strict scorer outcome (slice match and in-plane distance within 20 px). The multi-lesion variant ($n=14$) is omitted; no model produced a matched finding in the runs evaluated.}
\label{tab:longitudinal-behaviour} \\
\toprule
\multirow{2}{*}{\textbf{Model}} & \multirow{2}{*}{$n$} & \multirow{2}{*}{\textbf{Submitted \% $\uparrow$}} & \multirow{2}{*}{\textbf{Correct slice \% $\uparrow$}} & \multicolumn{2}{c}{\textbf{Slice diff $\downarrow$}} & \multicolumn{2}{c}{\textbf{3D mm $\downarrow$}} & \multirow{2}{*}{\textbf{Within 100 mm \% $\uparrow$}} & \multirow{2}{*}{\textbf{Hit \% $\uparrow$}} \\
\cmidrule(lr){5-6} \cmidrule(lr){7-8}
 & & & & med & p90 & med & p90 & & \\
\midrule
\endfirsthead
\multicolumn{10}{l}{\textit{Table~\ref{tab:longitudinal-behaviour} continued from previous page}} \\
\toprule
\multirow{2}{*}{\textbf{Model}} & \multirow{2}{*}{$n$} & \multirow{2}{*}{\textbf{Submitted \% $\uparrow$}} & \multirow{2}{*}{\textbf{Correct slice \% $\uparrow$}} & \multicolumn{2}{c}{\textbf{Slice diff $\downarrow$}} & \multicolumn{2}{c}{\textbf{3D mm $\downarrow$}} & \multirow{2}{*}{\textbf{Within 100 mm \% $\uparrow$}} & \multirow{2}{*}{\textbf{Hit \% $\uparrow$}} \\
\cmidrule(lr){5-6} \cmidrule(lr){7-8}
 & & & & med & p90 & med & p90 & & \\
\midrule
\endhead
\midrule
\multicolumn{10}{r}{\textit{continued on next page}} \\
\endfoot
\bottomrule
\endlastfoot
Claude Sonnet 4.6 & 90 & 92.2 & 1.1 & 35 & 92 & 134 & 215 & 20.5 & 0 \\
GPT-5.4 & 90 & 100 & 0 & 44.5 & 128 & 171 & 284 & 16.7 & 0 \\
GPT-5.4-nano & 90 & 94.4 & 0 & 67 & 121 & 166 & 259 & 12.9 & 0 \\
Gemini 3 Flash  & 90 & 86.7 & 0 & 42.5 & 98 & 166 & 267 & 16.7 & 0 \\
Gemini 3 Pro  & 90 & 95.6 & 3.3 & 43 & 112 & 167 & 267 & 16.3 & 0 \\
Qwen 3.5 & 90 & 91.1 & 1.1 & 31 & 93 & 146 & 232 & 19.5 & 0 \\
Gemma 4 & 90 & 100 & 0 & 36 & 99 & 179 & 308 & 15.6 & 0 \\
Ministral 3 (14B) & 90 & 98.9 & 0 & 45 & 98 & 146 & 208 & 18.0 & 0 \\
Mistral Large 3 & 90 & 87.8 & 0 & 41 & 92 & 146 & 208 & 19.0 & 0 \\
Kimi K2.5 & 90 & 98.9 & 2.2 & 35 & 107 & 152 & 247 & 18.0 & 0 \\
\end{longtable}
}

\section{Limitations and future work}
\label{apd:limitations}
\label{apd:future-work}
\textbf{Programmatic reference trajectories.} The Planning component compares each agent trajectory against a reference path generated programmatically from the task template at construction time. We acknowledge that this is our best estimate of a reasonable workflow rather than a radiologist-validated ground truth: the templates encode the steps that the task author judged necessary to solve the case, but they have not been reviewed against how a board-certified reader would actually drive the viewer. The empirical results offer indirect support for the construction. Planning and Execution scores are uniformly high on the easy and medium tiers (Table~\ref{tab:abra-results}), consistent with the reference path capturing a plausible solution route when the workflow is short and well-constrained. On the hard tier, Planning scores diverge across models and drop relative to the easier tiers, which is the regime in which the reference's coverage of valid alternative paths is weakest and where independent validation would be most informative. The natural next step is to collect real radiologist trajectories on the same tasks. We plan to extend the OHIF viewer to log the tool calls a human reader produces while completing each task, so that radiologist-derived references can replace, or be aggregated with, the programmatic ones per task. ABRA already provides the underlying components for this (a task definition shared between agent and human, a viewer that exposes the same tool surface to both, and scorers that consume tool-call traces); the remaining work is the data collection itself.

\textbf{Modality and dataset scope.} ABRA currently samples three TCIA datasets (LIDC-IDRI chest CT, Duke breast MRI, NLST longitudinal CT). The patient counts are modest, particularly for LIDC at 20 studies, and the modality spread excludes mammography, plain radiography, ultrasound, PET-CT, and nuclear medicine; the datasets also share a single US-clinical sourcing population. Two observations help put this scope in context. First, the small patient counts are partly offset by per-study annotation density: LIDC's multi-radiologist consensus contours and per-nodule attributes yield many independent tasks per study, so the task budget is not directly bottlenecked by patient count. Second, and more fundamentally, ABRA is designed to measure agentic capabilities through proxy tasks rather than to estimate clinical performance on population-level radiology. Its purpose is to expose capability gaps and provide a measurable signal that improvement can be tracked against; population-scale clinical accuracy is a different question, requiring a different data design, that we do not claim to answer. The three datasets already span volumetric detection (LIDC), structured cancer reporting (Duke), and longitudinal change (NLST), which we view as a reasonable diversity floor for the questions ABRA is built to ask. Expanding modality coverage is possible as future work on ABRA. Each new modality requires its own task generators and reference data, so this is genuine engineering work; on the architectural side, however, ABRA tasks can be constructed from any DICOM-based dataset with annotations.

\textbf{Radiology information system and FHIR integration.} Working radiologists routinely consume non-imaging context (prior reports, demographics, labs, pathology) when interpreting a study, and several natural extensions of ABRA require that context to be addressable. The current environment exposes the imaging archive (DICOM via Orthanc) and the viewer, but no structured electronic-health-record interface, so tasks that depend on correlating imaging findings with longitudinal clinical data are presently out of scope. Adding a FHIR endpoint to the controller's tool surface is a clean architectural extension: an embedded FHIR server can be added alongside Orthanc and exposed through additional tools without changing the agent loop or the scoring framework. One concrete direction is to integrate with the FHIR sandbox of MedAgentBench, so that radiology and EHR tasks share a common environment and the two benchmarks become composable. The bottleneck for this extension is data rather than architecture. Public datasets that pair DICOM imaging with structured longitudinal EHR data for the same patients are rare, and the partial pairings we are aware of require credentialed access through institutional data-use agreements. Even MedAgentBench, which is EHR-only with no imaging component, sourced its patient profiles from a single institution and relied on practising physicians to author its tasks, illustrating that the EHR substrate alone is non-trivial to assemble. ABRA's FHIR integration is therefore a natural future direction whose realisation is data-bound rather than design-bound.

\textbf{Multi-agent orchestration.} ABRA is presently a single-agent benchmark: one agent receives the task, drives the viewer and other tools through the controller's tool surface, and produces the final submission. The oracle variants of the annotation and BI-RADS reporting tasks partly address the question this single-agent design might otherwise leave open. By providing the agent with ground-truth upstream context (the lesion location for oracle annotation, the finding text for oracle reporting), these tasks mirror the situation in which an upstream perception or retrieval stage has succeeded and only the downstream subtask is being measured. The capability gap from upstream to downstream stages is therefore visible in the current results. What this design does not capture is error compounding across stages, the dynamic in which a perception subagent's miscalibrated output is consumed by a reporting subagent and propagates downstream. The direction in which specialised subagents collaborate, each on a small subtask of the workflow, is genuinely interesting, but evaluating it properly is out of scope for this work. Doing so on ABRA would require the scoring suite to attribute Planning, Execution, and Outcome to per-subagent contributions rather than to a single trajectory, which is a non-trivial change; it also depends on a task-decomposition convention that the agent literature has not yet converged on. Multi-agent orchestration is therefore a future direction whose realisation depends on both an extension of ABRA's scoring framework and broader progress on how subtasks should be defined.

\textbf{Task realism and clinical scope.} ABRA's current task set covers annotation, structured BI-RADS reporting, longitudinal new-lesion localisation, viewer control, and metadata querying. These are tractable proxies for radiologist behaviour rather than full reproductions of clinical work. A radiologist reading a chest CT in practice does not simply draw a circle around a nodule and submit; they reconcile the finding against the clinical indication, prior imaging, and the patient's broader history, then synthesise a free-text impression that serves the referring clinician. The proxies we score are deliberately narrow because narrow tasks have well-defined ground truth and admit automatic evaluation, but they leave large parts of the radiology workflow outside the benchmark. \citet{bluethgen2025agenticsystemsradiologydesign} sketch several higher-level agentic workflows that radiology agents would need to handle, including consistency checking on chest radiograph reports, end-to-end lung cancer screening reporting (scenario identification, longitudinal nodule tracking, Lung-RADS categorisation, and report drafting), agent-assisted resident tutoring, multidisciplinary team preparation, and follow-up scheduling. Each of these requires data and infrastructure that ABRA does not currently expose. Several further directions are also untouched in our task set: free-text full-report synthesis with paragraph-level scoring against a reference report, radiation treatment planning workflows, and similar workflow-level tasks that integrate imaging interpretation with downstream clinical actions. We do not claim that strong scores on ABRA imply readiness for any of these higher-level applications. The benchmark is positioned as a measurement of the perception and orchestration substrate that such applications would build on; closing the realism gap between this substrate and the workflows above is the principal direction of further work, and depends on the same data and scoring extensions discussed in the FHIR and multi-agent paragraphs above.

%%%%%%%%%%%%%%%%%%%%%%%%%%%%%%%%%%%%%%%%%%%%%%%%%%%%%%%%%%%%
\section{Total token budget per model}

\begin{table}[h]
\centering
\small
\caption{Per-model token consumption and wall-clock cost across the full benchmark. Tokens are summed across tasks in each difficulty bucket as reported by each provider's billing-side counters. \emph{Runtime} is the sum of per-task durations, which is independent of how many tasks were run in parallel during the benchmark.}
\label{tab:execution-budget}
\begin{tabular}{l r r r r r}
\toprule
\multirow{2}{*}{\textbf{Model}} & \multicolumn{4}{c}{\textbf{Tokens (M)}} & \multirow{2}{*}{\textbf{Runtime (h)}} \\
\cmidrule(lr){2-5}
 & Easy & Medium & Hard & \textbf{Total} & \\
\midrule
Claude Sonnet 4.6        & 0.5 & 5.8 & 115.2 & \textbf{121.4} & 14.0 \\
GPT-5.4                  & 0.5 & 4.4 &  29.7 & \textbf{ 34.7} &  4.2 \\
GPT-5.4-nano             & 0.5 & 4.9 &  34.7 & \textbf{ 40.2} &  3.6 \\
Gemini 3 Flash  & 1.3 & 8.3 &  18.4 & \textbf{ 28.0} &  2.9 \\
Gemini 3 Pro    & 1.4 & 5.6 &  53.1 & \textbf{ 60.2} & 16.0 \\
Qwen 3.5                 & 1.0 & 7.8 &  29.3 & \textbf{ 38.0} & 17.9 \\
Gemma 4                  & 0.6 & 4.5 &  11.6 & \textbf{ 16.6} & 16.4 \\
Ministral 3 (14B)        & 0.9 & 5.8 &  11.9 & \textbf{ 18.5} &  3.5 \\
Mistral Large 3          & 0.8 & 6.8 &  14.6 & \textbf{ 22.2} &  6.5 \\
Kimi K2.5                & 1.1 & 8.5 &  25.3 & \textbf{ 34.8} & 17.2 \\
\bottomrule
\end{tabular}
\end{table}

\section{Per-task-type token usage and latency}
\label{apd:token-usage}

Table~\ref{tab:token-usage} reports mean per-task token usage, turn count, and wall-clock duration broken down by model and task type.

{\footnotesize
\setlength{\tabcolsep}{3pt}
\begin{longtable}{@{}l l r r r r@{}}
\caption{Per-model, per-task-type token usage and latency, averaged over all tasks of the type. \emph{Input} and \emph{Output} are mean tokens per task. \emph{Turns} is the mean number of agent turns per episode. \emph{Dur.} is the mean wall-clock duration per task in seconds. Token counts are reported with a $\text{k}$ suffix for $\geq 10^3$ and a $\text{M}$ suffix for $\geq 10^6$.}
\label{tab:token-usage} \\
\toprule
\textbf{Model} & \textbf{Task type} & \textbf{Input} & \textbf{Output} & \textbf{Turns} & \textbf{Dur.\ (s)} \\
\midrule
\endfirsthead
\multicolumn{6}{l}{\textit{Table~\ref{tab:token-usage} continued from previous page}} \\
\toprule
\textbf{Model} & \textbf{Task type} & \textbf{Input} & \textbf{Output} & \textbf{Turns} & \textbf{Dur.\ (s)} \\
\midrule
\endhead
\midrule
\multicolumn{6}{r}{\textit{continued on next page}} \\
\endfoot
\bottomrule
\endlastfoot
Claude Sonnet 4.6 & \texttt{viewer\_control}        & 3.0k  & 312   & 2.0  & 8.2   \\
                  & \texttt{metadata\_qa}           & 1.6k  & 259   & 2.1  & 6.4   \\
                  & \texttt{vision\_probe}          & 1.1k  & 101   & 1.0  & 10.2  \\
                  & \texttt{annotation}             & 7     & 1.3k  & 5.4  & 29.1  \\
                  & \texttt{oracle\_annotation}     & 10    & 3.1k  & 7.7  & 41.3  \\
                  & \texttt{oracle\_birads\_report} & 4     & 372   & 2.0  & 7.3   \\
                  & \texttt{longitudinal}           & 28    & 15.5k & 25.6 & 327.5 \\
                  & \texttt{birads\_report}         & 10    & 3.7k  & 8.5  & 150.2 \\
\addlinespace
GPT-5.4           & \texttt{viewer\_control}        & 2.3k  & 52    & 2.6  & 3.4   \\
                  & \texttt{metadata\_qa}           & 2.9k  & 104   & 2.0  & 3.1   \\
                  & \texttt{vision\_probe}          & 1.0k  & 17    & 1.0  & 2.2   \\
                  & \texttt{annotation}             & 13.4k & 351   & 4.6  & 11.5  \\
                  & \texttt{oracle\_annotation}     & 17.1k & 1.8k  & 5.3  & 23.5  \\
                  & \texttt{oracle\_birads\_report} & 3.1k  & 119   & 2.0  & 3.4   \\
                  & \texttt{longitudinal}           & 146k  & 3.8k  & 9.4  & 93.0  \\
                  & \texttt{birads\_report}         & 36.9k & 1.9k  & 4.3  & 35.7  \\
\addlinespace
GPT-5.4-nano      & \texttt{viewer\_control}        & 1.8k  & 73    & 2.2  & 4.0   \\
                  & \texttt{metadata\_qa}           & 3.1k  & 106   & 2.0  & 4.4   \\
                  & \texttt{vision\_probe}          & 1.0k  & 17    & 1.0  & 2.5   \\
                  & \texttt{annotation}             & 13.9k & 427   & 4.6  & 10.5  \\
                  & \texttt{oracle\_annotation}     & 22.9k & 1.5k  & 6.7  & 16.6  \\
                  & \texttt{oracle\_birads\_report} & 3.1k  & 124   & 2.0  & 2.6   \\
                  & \texttt{longitudinal}           & 174k  & 1.6k  & 20.8 & 75.2  \\
                  & \texttt{birads\_report}         & 24.9k & 824   & 5.5  & 36.8  \\
\addlinespace
Gemini 3 Flash  & \texttt{viewer\_control}        & 11.4k & 213   & 7.0  & 11.1  \\
                  & \texttt{metadata\_qa}           & 4.9k  & 238   & 2.7  & 4.9   \\
                  & \texttt{vision\_probe}          & 1.5k  & 16    & 1.0  & 4.5   \\
                  & \texttt{annotation}             & 17.9k & 499   & 4.9  & 13.5  \\
                  & \texttt{oracle\_annotation}     & 59.6k & 2.9k  & 8.5  & 25.9  \\
                  & \texttt{oracle\_birads\_report} & 4.8k  & 232   & 2.9  & 4.8   \\
                  & \texttt{longitudinal}           & 103k  & 1.7k  & 9.9  & 41.3  \\
                  & \texttt{birads\_report}         & 31.4k & 822   & 5.2  & 20.3  \\
\addlinespace
Gemini 3 Pro  & \texttt{viewer\_control}        & 12.3k & 977   & 7.8  & 28.8  \\
                  & \texttt{metadata\_qa}           & 3.2k  & 839   & 2.2  & 11.7  \\
                  & \texttt{vision\_probe}          & 1.5k  & 1.4k  & 1.0  & 16.0  \\
                  & \texttt{annotation}             & 16.1k & 12.3k & 4.5  & 103.1 \\
                  & \texttt{oracle\_annotation}     & 13.4k & 3.9k  & 4.9  & 62.4  \\
                  & \texttt{oracle\_birads\_report} & 4.1k  & 769   & 2.4  & 16.6  \\
                  & \texttt{longitudinal}           & 273k  & 29.4k & 8.7  & 270.4 \\
                  & \texttt{birads\_report}         & 114k  & 9.4k  & 6.3  & 115.7 \\
\addlinespace
Qwen 3.5          & \texttt{viewer\_control}        & 3.4k  & 478   & 2.4  & 14.2  \\
                  & \texttt{metadata\_qa}           & 5.0k  & 653   & 2.0  & 11.0  \\
                  & \texttt{vision\_probe}          & 1.2k  & 1.3k  & 1.0  & 32.7  \\
                  & \texttt{annotation}             & 29.8k & 2.0k  & 5.8  & 56.6  \\
                  & \texttt{oracle\_annotation}     & 57.7k & 4.4k  & 7.9  & 114.8 \\
                  & \texttt{oracle\_birads\_report} & 5.4k  & 567   & 2.0  & 10.3  \\
                  & \texttt{longitudinal}           & 211k  & 8.8k  & 15.1 & 351.7 \\
                  & \texttt{birads\_report}         & 122k  & 6.0k  & 13.4 & 177.5 \\
\addlinespace
Gemma 4           & \texttt{viewer\_control}        & 2.0k  & 64    & 2.0  & 8.3   \\
                  & \texttt{metadata\_qa}           & 3.9k  & 141   & 2.0  & 9.0   \\
                  & \texttt{vision\_probe}          & 731   & 15    & 1.0  & 8.0   \\
                  & \texttt{annotation}             & 17.1k & 450   & 4.1  & 98.7  \\
                  & \texttt{oracle\_annotation}     & 33.6k & 2.3k  & 7.5  & 162.0 \\
                  & \texttt{oracle\_birads\_report} & 4.1k  & 141   & 2.0  & 23.6  \\
                  & \texttt{longitudinal}           & 94.9k & 1.6k  & 11.3 & 283.7 \\
                  & \texttt{birads\_report}         & 30.2k & 622   & 5.2  & 85.1  \\
\addlinespace
Ministral 3 (14B) & \texttt{viewer\_control}        & 2.4k  & 187   & 2.3  & 3.9   \\
                  & \texttt{metadata\_qa}           & 6.5k  & 174   & 2.5  & 4.1   \\
                  & \texttt{vision\_probe}          & 1.1k  & 15    & 1.0  & 5.4   \\
                  & \texttt{annotation}             & 23.4k & 561   & 5.3  & 17.2  \\
                  & \texttt{oracle\_annotation}     & 34.3k & 2.9k  & 6.1  & 31.3  \\
                  & \texttt{oracle\_birads\_report} & 4.5k  & 141   & 2.0  & 2.9   \\
                  & \texttt{longitudinal}           & 76.8k & 1.6k  & 11.7 & 45.7  \\
                  & \texttt{birads\_report}         & 71.9k & 2.4k  & 9.0  & 45.3  \\
\addlinespace
Mistral Large 3   & \texttt{viewer\_control}        & 4.5k  & 121   & 3.4  & 13.1  \\
                  & \texttt{metadata\_qa}           & 4.3k  & 132   & 2.0  & 7.8   \\
                  & \texttt{vision\_probe}          & 1.1k  & 11    & 1.0  & 5.7   \\
                  & \texttt{annotation}             & 24.5k & 551   & 5.5  & 24.5  \\
                  & \texttt{oracle\_annotation}     & 51.3k & 2.4k  & 7.9  & 44.2  \\
                  & \texttt{oracle\_birads\_report} & 4.5k  & 169   & 2.0  & 6.6   \\
                  & \texttt{longitudinal}           & 117k  & 1.9k  & 15.1 & 125.5 \\
                  & \texttt{birads\_report}         & 43.0k & 1.7k  & 7.7  & 42.9  \\
\addlinespace
Kimi K2.5         & \texttt{viewer\_control}        & 8.7k  & 643   & 6.8  & 21.5  \\
                  & \texttt{metadata\_qa}           & 3.2k  & 451   & 2.0  & 10.5  \\
                  & \texttt{vision\_probe}          & 1.2k  & 740   & 1.0  & 17.3  \\
                  & \texttt{annotation}             & 19.3k & 1.2k  & 5.8  & 45.3  \\
                  & \texttt{oracle\_annotation}     & 101k  & 3.5k  & 12.8 & 58.7  \\
                  & \texttt{oracle\_birads\_report} & 2.8k  & 373   & 2.0  & 7.1   \\
                  & \texttt{longitudinal}           & 190k  & 4.3k  & 11.8 & 379.8 \\
                  & \texttt{birads\_report}         & 98.8k & 3.5k  & 11.6 & 167.3 \\
\end{longtable}
}

%%%%%%%%%%%%%%%%%%%%%%%%%%%%%%%%%%%%%%%%%%%%%%%%%%%%%%%%%%%%
\newpage

\end{document}